\pdfoutput=1
\documentclass[11pt]{article}

\usepackage[final]{acl}

\usepackage{times}
\usepackage{latexsym}

\usepackage[T1]{fontenc}
\usepackage[utf8]{inputenc}

\usepackage{microtype}

\usepackage{inconsolata}

\usepackage{graphicx}

\usepackage{multirow} % Required for multirows
\usepackage{amsmath,amssymb,amsfonts}%
\usepackage{wasysym}
\usepackage{amsthm}%
\usepackage{mathrsfs}%
\usepackage{xcolor}%
\usepackage{textcomp}%
\usepackage{manyfoot}%
\usepackage{booktabs}%
\usepackage{algorithm}%
\usepackage{algorithmicx}%
\usepackage{algpseudocode}%
\usepackage{listings}%
\usepackage{array}
\usepackage{makecell}

\usepackage{enumitem}
\usepackage{pifont} % For checkmark and cross
\usepackage{booktabs}
\usepackage{subfig}
\usepackage{hyperref}
\usepackage{ragged2e}
\newcommand{\wt}{\textcolor[RGB]{0,0,0}}

\newcommand{\ourbenchmark}{{\fontfamily{ppl}\selectfont Asclepius}}

\newcommand{\cmark}{\ding{51}}%
\newcommand{\xmark}{\ding{55}}%
\newcommand{\yes}{\textcolor{green}{\cmark}}
\newcommand{\no}{\textcolor{red}{\xmark}}

\title{\ourbenchmark: A Spectrum Evaluation Benchmark for \\ Medical Multi-Modal Large Language Models}

\author{Jie Liu $^{1*}$,
        Wenxuan Wang$^2$\thanks{J. Liu ($\spadesuit$ $\clubsuit$ $\diamondsuit$), W. Wang ($\heartsuit$ $\clubsuit$) make equal contribution. $\spadesuit$: Conceptualization; $\heartsuit$: Implementation; $\clubsuit$: Writing; $\diamondsuit$: Visualization.},
	Yihang Su$^2$,
	Jingyuan Huan$^2$,
    Wenting Chen $^1$,
    \textbf{Yudi Zhang} $^3$,\\
    \textbf{Cheng-Yi Li}$^{4,5}$,
    \textbf{Kao-Jung Chang}$^{4,5}$,
    \textbf{Xiaohan Xing}$^{6}$,
    \textbf{Linlin Shen}$^3$,
    \textbf{Michael R. Lyu}$^2$ \\[4mm]
    $^1$The City University of Hong Kong~$^2$The Chinese University of Hong Kong\\ [0.5mm]
    $^3$Shenzhen University~$^4$National Yang Ming Chiao Tung University\\ [0.5mm]
    $^5$Taipei Veterans General Hospital~$^6$Stanford University\\ [0.5mm]
    {\small Project Page:~\href{https://asclepius-med.github.io/}{https://asclepius-med.github.io/}}
    }

\begin{document}
\maketitle
\begin{abstract}
The significant breakthroughs of Medical Multi-Modal Large Language Models (Med-MLLMs) renovate modern healthcare with robust information synthesis and medical decision support. However, these models are often evaluated on benchmarks that are unsuitable for the Med-MLLMs due to the complexity of real-world diagnostics across diverse specialties.
To address this gap, we introduce \ourbenchmark, a novel Med-MLLM benchmark that comprehensively assesses Med-MLLMs in terms of: distinct medical specialties (cardiovascular, gastroenterology, etc.) and different diagnostic capacities (perception, disease analysis, etc.). Grounded in 3 proposed core principles, \ourbenchmark~ensures a comprehensive evaluation by encompassing 15 medical specialties, stratifying into 3 main categories and 8 sub-categories of clinical tasks, and exempting overlap with existing VQA dataset. We further provide an in-depth analysis of 6 Med-MLLMs and compare them with 3 human specialists, providing insights into their competencies and limitations in various medical contexts. Our work not only advances the understanding of Med-MLLMs' capabilities but also sets a precedent for future evaluations and the safe deployment of these models in clinical environments.
\end{abstract}

\section{Introduction}

The advent of Multi-Modal Large Language Models (MLLMs), such as GPT-4V \citep{openai2023gpt4v}, Gemini \citep{team2023gemini}, LLaVA \citep{liu2023visual}, and MiniGPT-4 \citep{zhu2023minigpt}, represents a significant stride towards artificial general intelligence due to their exceptional proficiency in tackling intricate tasks. These advancements have not only expanded the capabilities of models {in natural scenes} but have also paved the way for specialized enhancement in healthcare, as seen with the emergence of recent Medical Multi-modal Large Language Models (Med-MLLMs) \citep{moor2023foundation,liu2023medical,lee2023cxr,zhang2023pmc}.

\begin{figure}[t]
\centering
	{\includegraphics[width=0.90\linewidth]{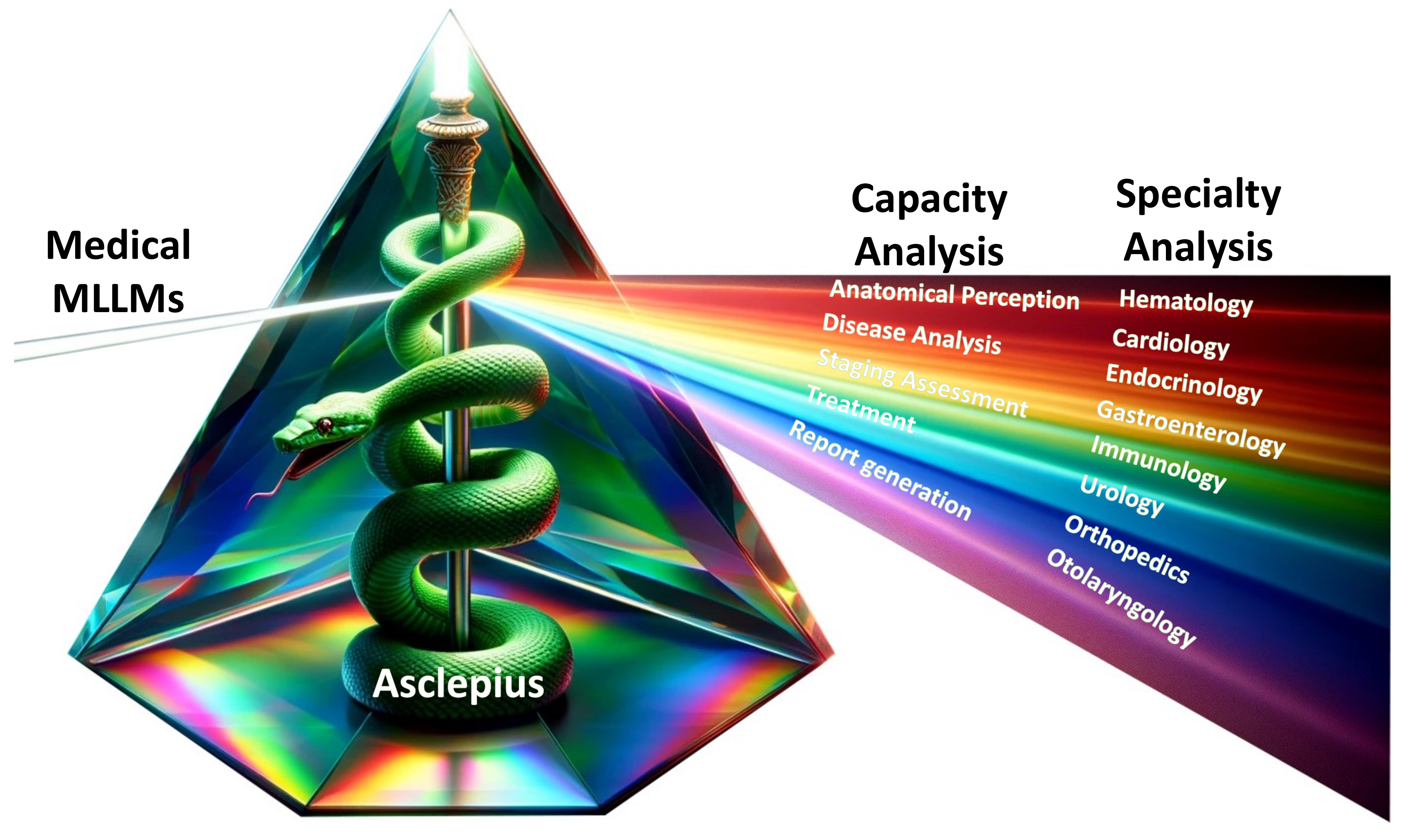}}
 \vspace{-0.1cm}
	\caption{\textbf{\ourbenchmark,}{ a spectrum evaluation benchmark for Med-MLLMs, analyzes models on the capacity dimension with 8 clinical tasks and the specialty dimension with 15 medical specialties.} }
	% Grounded in three core principles, \ourbenchmark~ensures a comprehensive evaluation by encompassing 15 medical specialties, stratifying into 3 main categories and 8 sub-categories of clinical tasks, and avoiding data contamination by using novel datasets. 
	\label{fig:teaser}
 \vspace{-0.6cm}
\end{figure}

\begin{table*}[t]
	\centering
	\caption{\textbf{Comparison of Med-MLLMs' Datasets.} The \ourbenchmark~is categorized by both medical specialty and capability, encompassing 15 major specialties and 8 core competencies, and includes human specialists evaluation scores.}
 \vspace{-0.2cm}
 \scalebox{0.80}{
	\begin{tabular}{
	 		ll
	 		>{\centering\arraybackslash}p{2cm} % Column 3 width with auto line break
	 		>{\centering\arraybackslash}p{1.5cm} % Column 4 width with auto line break
	 		>{\centering\arraybackslash}p{2cm} % Column 5 width with auto line break
	 		>{\centering\arraybackslash}p{2cm} % Column 6 width with auto line break
	 		>{\centering\arraybackslash}p{2cm} % Column 6 width with auto line break
	 	}
		\toprule
		\multirow{2}{*}{Name} & \multirow{2}{*}{Modality} & \multirow{2}{*}{\makecell{Body parts \\and organs}} & \multicolumn{2}{c}{Division} & \multirow{2}{*}{Original} & \multirow{2}{*}{\makecell{Human \\ Evaluation}}\\
		\cmidrule(r){4-5}
		& & & Specialty & Capacity & \\
		\midrule
		ROCO \citep{pelka2018radiology} & Radiology & / & \no & \no & \yes & \no \\
		VQA-RAD \citep{lau2018dataset} & Radiology & 3 & \no & \no & \yes & \no \\
		SLAKE \citep{liu2021slake} & Radiology & 5 & \no & \no & \yes & \no \\
		PathVQA \citep{he2021towards} & Pathology & / & \no & \no & \yes &\no \\
		MedMD \citep{wu2023towards} & Radiology & 17 & \no & \no & \no & \no \\
		PMC-VQA \citep{zhang2023pmc} & Multi-Modality & / & \no & \no & \no &\no \\
		\ourbenchmark & Multi-Modality & 79 & \yes & \yes & \yes &\yes \\
		\bottomrule
	\end{tabular}}
	\label{tab:medical_datasets}
 \vspace{-0.4cm}
\end{table*}

Despite the promising advancements, the evaluation of these models {relies} predominantly on a limited set of samples \citep{wu2023can,li2023comprehensive,zhou2023skingpt,moor2023foundation}, which provides {an} incomplete picture of their capabilities. Current {medical} benchmarks, originally designed for traditional learning models, fall short in measuring the sophisticated capabilities of Med-MLLMs \citep{moor2023med,li2023llava}. This misalignment highlights the necessity for comprehensive benchmarks that can systematically assess diverse perspectives of Med-MLLM. To this end, we propose a novel benchmark \ourbenchmark~that is akin to analyzing the spectrum of light with a prism, as illustrated in Figure \ref{fig:teaser}.

Developing such a benchmark is challenging, especially in the medical field, due to the variation in practical expertise across different domains \citep{wang2023medfmc}. For instance, a cardiologist may require a referral to a gastroenterologist when encountering conditions outside their primary domain of expertise due to the distinct specializations within each field \citep{forrest2000coordination,forrest2001prevalence}. Traditional model evaluation methods often overlook the need for specialty-specific assessments, resulting in limited applicability and reliability in specialized medical contexts. This drives our approach to consider the variation in expertise across different medical specialties and to assess Med-MLLMs' abilities in specialization-specific knowledge.
Furthermore, the complex clinical decision-making processes, which involve perception, cognition, and reasoning \citep{lyman2023perception,liberatore2008analytic,patel2002emerging,kassirer1978clinical}, present a second challenge. Med-MLLMs {should} have a diverse set of capabilities, such as interpreting medical imagery and understanding pathophysiology, to mimic these processes. Current evaluation frameworks \citep{lau2018dataset,pelka2018radiology,liu2021slake}, which often focus on tasks like general radiology VQA, fail to precisely capture clinical performance. Thus, three core principles is proposed to address these concerns.

%Thirdly, the evaluation data from the currently available dataset is easily cross-contaminated with training data for Med-MLLMs, thereby leading to potential data leakage. In an era where many models are trained on vast swaths of publicly available data, there is a significant risk of evaluating models on data they have previously encountered \citep{magar2022data}. The phenomenon of data leakage may artificially inflate the accuracy of these medical models, leading to misleadingly high performance \citep{deng2023benchmark}. Thus, the benchmark of Med-MLLMs should consider this risk by avoiding the use of existing publicly available datasets as much as possible.

\begin{figure*}[t]
\centering
	{\includegraphics[width=0.92\linewidth]{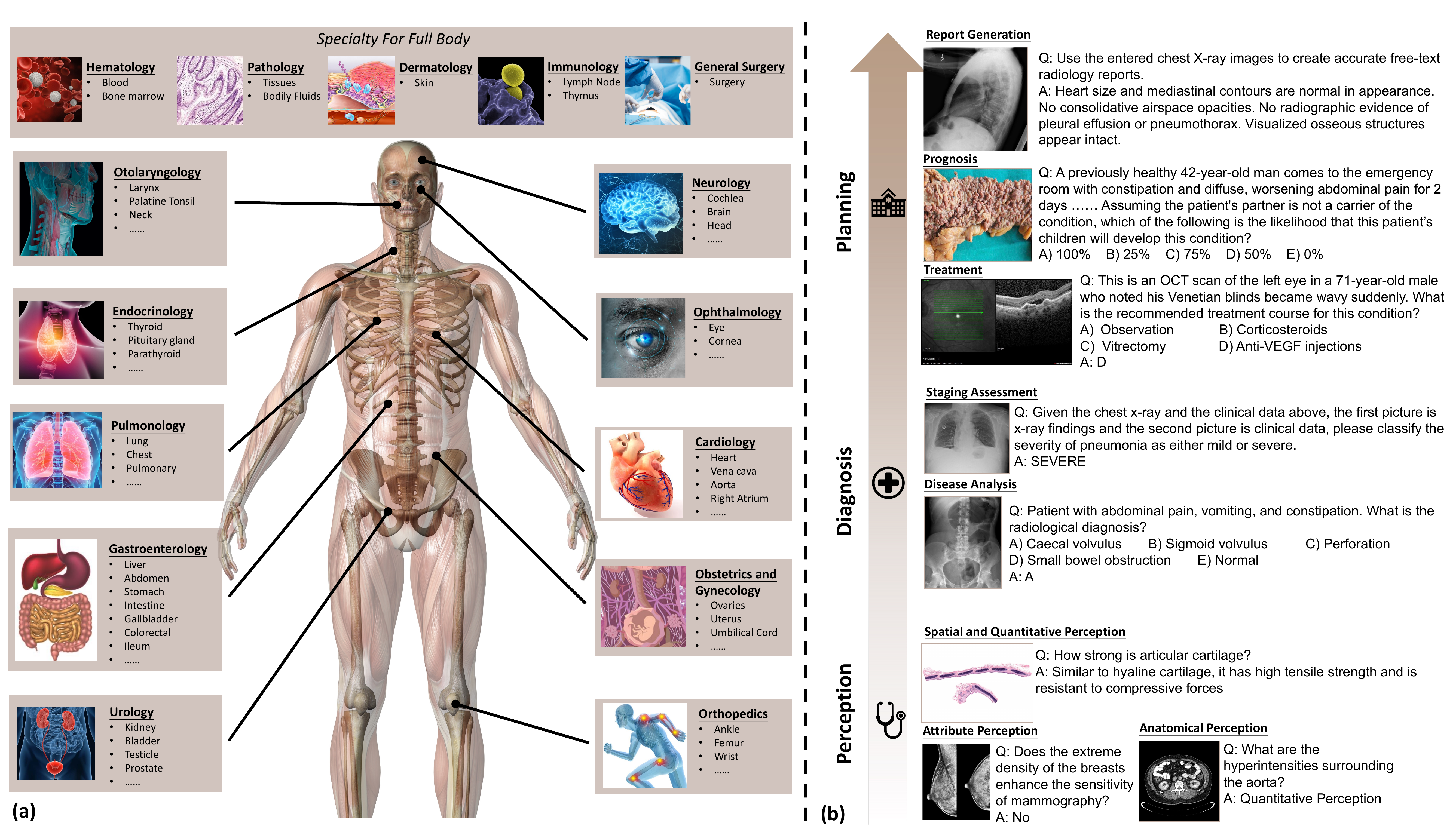}}
 \vspace{-0.1cm}
	\caption{\textbf{\ourbenchmark~Overview.}
		(a) Involves 15 specialties and 79 body parts and organs in total, representing the critical component of the healthcare system.
		(b) Shows examples for 8 distinct capacities, offering a multifaceted evaluation of Med-MLLMs.}
	\label{fig:specialty}
 \vspace{-0.6cm}
\end{figure*}

\noindent\textbf{The philosophy to create \ourbenchmark} (1) \textit{Multi-Specialty Coverage}: Our benchmark is meticulously designed to encompass a spectrum of 15 medical specialties. By systematically including questions from various specialties, such as cardiology, neurology, hematology, and endocrinology, the benchmark can evaluate the performance of Med-MLLMs in different medical domains. (2) \textit{Multi-Dimensional Capacity}: Acknowledging the intricacies inherent in medical problem-solving, our benchmark is meticulously designed to evaluate a spectrum, divided into 3 main categories and 8 sub-categories. (3) \textit{Original and Blindness}: The questions in MedVQABench are sourced from contemporary educational materials, medical examinations, and visual datasets, rather than integrating previously existing multiple VQA datasets, ensuring the originality of our benchmark. Moreover, we have developed a website that allows submission and server-side evaluation of results to ensure integrity and fairness in the evaluation process.

In summary, our main contributions are summarized as {follows}:
\begin{itemize}[leftmargin=10pt,itemsep=0pt]
	\item Systematically-Constructed Dataset: Our study introduces a meticulously crafted dataset designed to evaluate Med-MLLMs. This dataset encompasses a comprehensive range of 15 medical specialties, targeting 79 distinct body parts and organs. Furthermore, it is stratified into 3 main categories and 8 sub-categories, each corresponding to specific capacities within the medical domain.
	\item Comprehensive Benchmarking: Our study establishes a rigorous benchmark for the comprehensive assessment of 6 Med-MLLMs. In addition, 3 human doctors from varied specialties and levels of experience, ranging from junior to senior, answer the question to evaluate the human performance in this benchmark. This benchmark enables a direct comparison between Med-MLLMs and human specialists, providing valuable insights into the current state of AI in healthcare.
	\item Analysis and Observations: We provide several insights with corresponding suggestions (\S \ref{sec:suggestion}) based on evaluation results, shedding light on their strengths and weaknesses for Med-MLLMs.
\end{itemize}

\section{Related Work}
\subsection{Medical Multi-Modal Large Language Models (Med-MLLMs)}
The field of medicine frequently engages with diverse data modalities, including but not limited to text, computed tomography (CT) scans, dermoscopy images, and histopathological slides. In order to effectively replicate the complex decision-making processes of healthcare professionals, Medical Multi-Modal Large Language Models (Med-MLLMs) have been developed. Initial research efforts in this area have focused on the fusion of text with single medical imaging modalities \citep{liu2023medical,lee2023cxr,zhou2023skingpt,thawkar2023xraygpt,wu2023towards}. These contributions laid the groundwork for subsequent advances in the field. Progressing further, recent studies have aimed to amalgamate a broader range of modalities \citep{belyaeva2023multimodal,zhang2023biomedgpt,zhang2023pmc,li2023llava,zhang2023m3exam}. 
In \ourbenchmark, we thoroughly assess the capabilities of these models, particularly in terms of their performance in various specialties and capacities.

%\citep{liu2023medical} for diagnosing COVID-19 based on radiography
%\citep{lee2023cxr} for Interpreting Chest X-ray Images
%\citep{moor2023foundation} generalist medical AI
%\citep{belyaeva2023multimodal} for individual-specific data

\subsection{Benchmark for Med-MLLMs}
In the rapidly evolving domain of Med-MLLMs, the development of benchmarks for evaluating these models is of paramount concern. Recent works \citep{zhang2023biomedgpt,belyaeva2023multimodal,li2023llava,lu2023foundational,liu2023medical} have endeavored to aggregate data from a variety of publicly available sources \citep{subramanian2020medicat,yang2023medmnist,irvin2019chexpert,johnson2019mimic,he2021towards,lau2018dataset,liu2021slake} to create larger and more comprehensive datasets. Subsequently, ChatGPT is employed to assist in filtering the aggregated data to ensure quality control. 
For instance, the MedMD \citep{wu2023towards} collected data from existing visual-language medical datasets, such as MIMIC-CXR \citep{johnson2019mimic} and PMC-OA \citep{lin2023pmc}, within the radiology domain. Similarly, the PMC-VQA Dataset \citep{zhang2023pmc} utilizes ChatGPT to create a Visual Question Answering (VQA) dataset based on image-text pairings from PMC-OA \citep{lin2023pmc}. 
The recently proposed OmnimedVQA \cite{hu2024omnimedvqa} and GMAI-MMBench~\citep{chen2024gmai} also evaluates Med-MLLMs with single and multiple choice questions, which has been proved to be limited and biased~\cite{li2024can}, and cannot assess the models' generative capabilities and align with real-world settings.
We compare \ourbenchmark~with existing works in Table \ref{tab:medical_datasets}. Different from previous works, in this paper, we propose a novel Med-MLLMs benchmark \ourbenchmark, which is mainly built based on existing medical textbooks and medical image datasets \wt{with various question types}. We aim to establish a new standard for Med-MLLM evaluation that upholds the integrity of the assessment and delivers accurate reflection of a model's true capabilities in clinical environment.

\section{Asclepius Benchmark}%\ourbenchmark~Benchmark}
Our \ourbenchmark~benchmark contains 3,232 original multi-modal questions, with {a} spectrum of 15 medical specialties and 8 capacities evaluation. {This section is organized as follows}: In Sec. \ref{sec:specialty} and Sec. \ref{sec:capacity}, we discuss the design philosophy behind \ourbenchmark~ and then present the specialty and capacity splitting. In Sec. \ref{sec:source_statistics}, we details the question construction and provide statistics of \ourbenchmark.
%elaborate on the following organization of details
%briefly
%splitting
%we introduce how we construct the \ourbenchmark~questions, and provide statistics of \ourbenchmark.
\begin{figure}[t]
\centering
	{\includegraphics[width=0.77\linewidth]{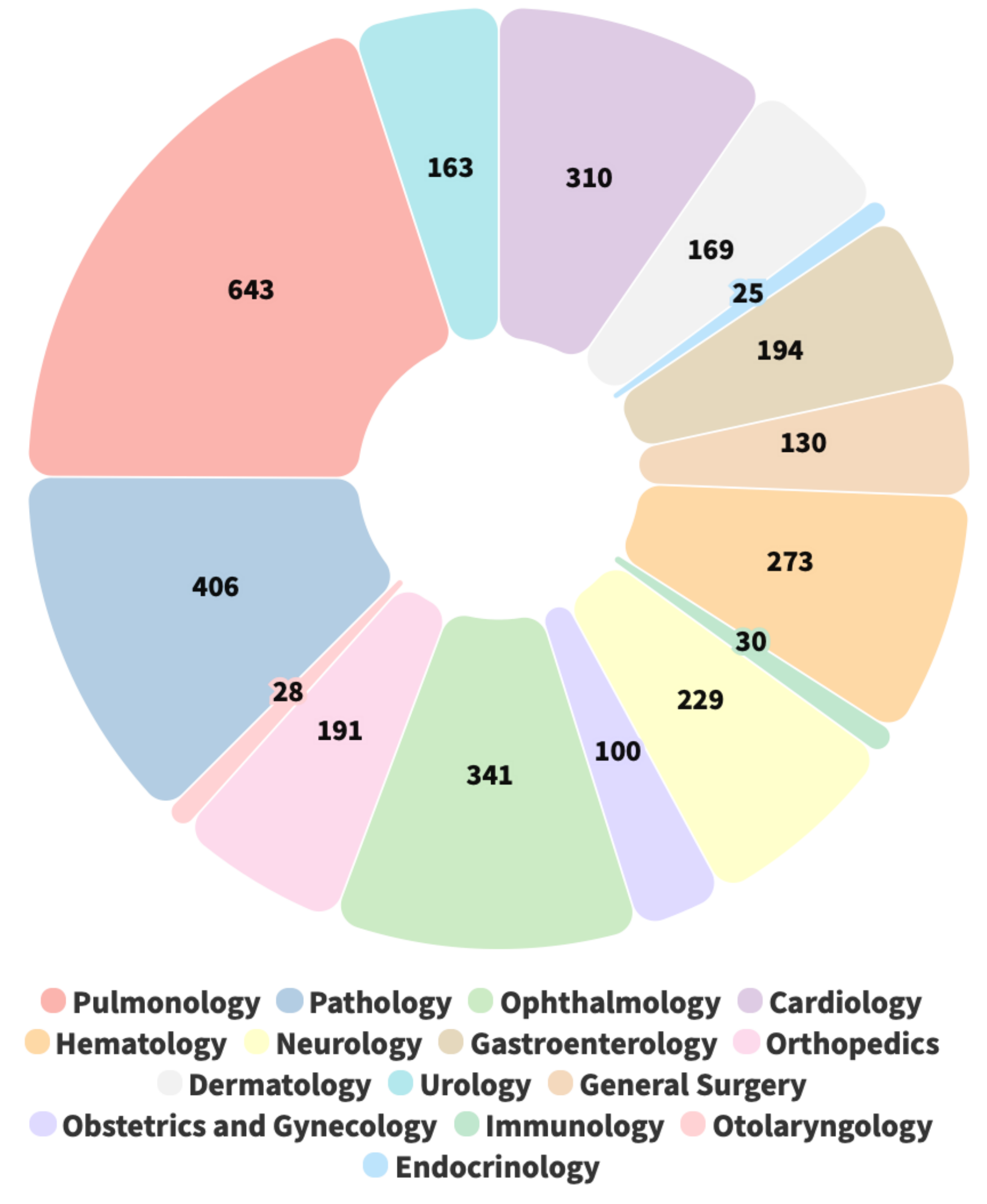}}
 \vspace{-0.1cm}
	\caption{\textbf{Data Statistics for Specialty}. Currently, \ourbenchmark~incorporates 15 specialties with 3,232 multi-modal questions.}
	\label{fig:specialty_sta}
 \vspace{-0.5cm}
\end{figure}

\subsection{Multi-Specialty Coverage}
\label{sec:specialty}
Medical education is characterized by significant disparities in knowledge across different specialties, necessitating rigorous training for medical students within their chosen fields \citep{ledford2022medical,davis2009help}. 
%This specialized training is fundamental to patient care, ensuring that medical professionals are equipped with the expertise required to deliver optimal treatment. 
Analogously, Med-MLLMs developed to support clinical decision-making {should} exhibit a comparable depth of knowledge in their respective specialties to be effective.
To this end, we include a diverse array of specialties. For specialties concerning specific organs, we incorporate \textit{Cardiology, Endocrinology, Obstetrics and Gynecology, Gastroenterology, Urology, Orthopedics, Neurology, Otolaryngology, Pulmonology}, and \textit{Ophthalmology}. In addition, for specialties encompassing full-body considerations, we include \textit{Hematology, Pathology, Dermatology, Immunology}, and \textit{General Surgery}. This diversified inclusion ensures that the Med-MLLMs reflect not only technological innovation but also practical utility within the routine operations of clinical practice.
%, thereby improving their reliability and applicability in actual healthcare settings.

To provide a comprehensive view of the Med-MLLMs across these specialties, Figure \ref{fig:specialty} (a) offers an overview of the 15 specialties assessed in \ourbenchmark. Each specialty represents a critical component of the healthcare system, addressing unique health concerns within its area of expertise. For additional information on the specialties included in this study, please refer to Appendix~\ref{sec:append_specialty}.%Supplementary Materials.

%Our benchmark dataset is strategically divided into two primary dimensions: Perception and Cognition. Within the Perception dimension, we have two subclasses: Coarse-Grained and Fine-Grained. The Coarse-Grained subclass encompasses a variety of questions assessing Existence, Location, Size, Color, Shape, Modality, Plane, Organ system, Position reasoning, and Counting capabilities. The Fine-Grained subclass is focused on more detailed evaluations such as Abnormality, Appearance, Attribute, and Instrument-related questions. The Cognition dimension is tailored to evaluate higher-order medical reasoning, characterized by Report generation and interpretation of Tables, Charts, and Documents. Each subclass incorporates questions involving at least one medical imaging modality, including MRI, X-ray, Ultrasound, Histopathology, and CT, ensuring a comprehensive assessment across different diagnostic tools. 

\begin{figure}[t]

\centering
	{\includegraphics[width=0.87\linewidth]{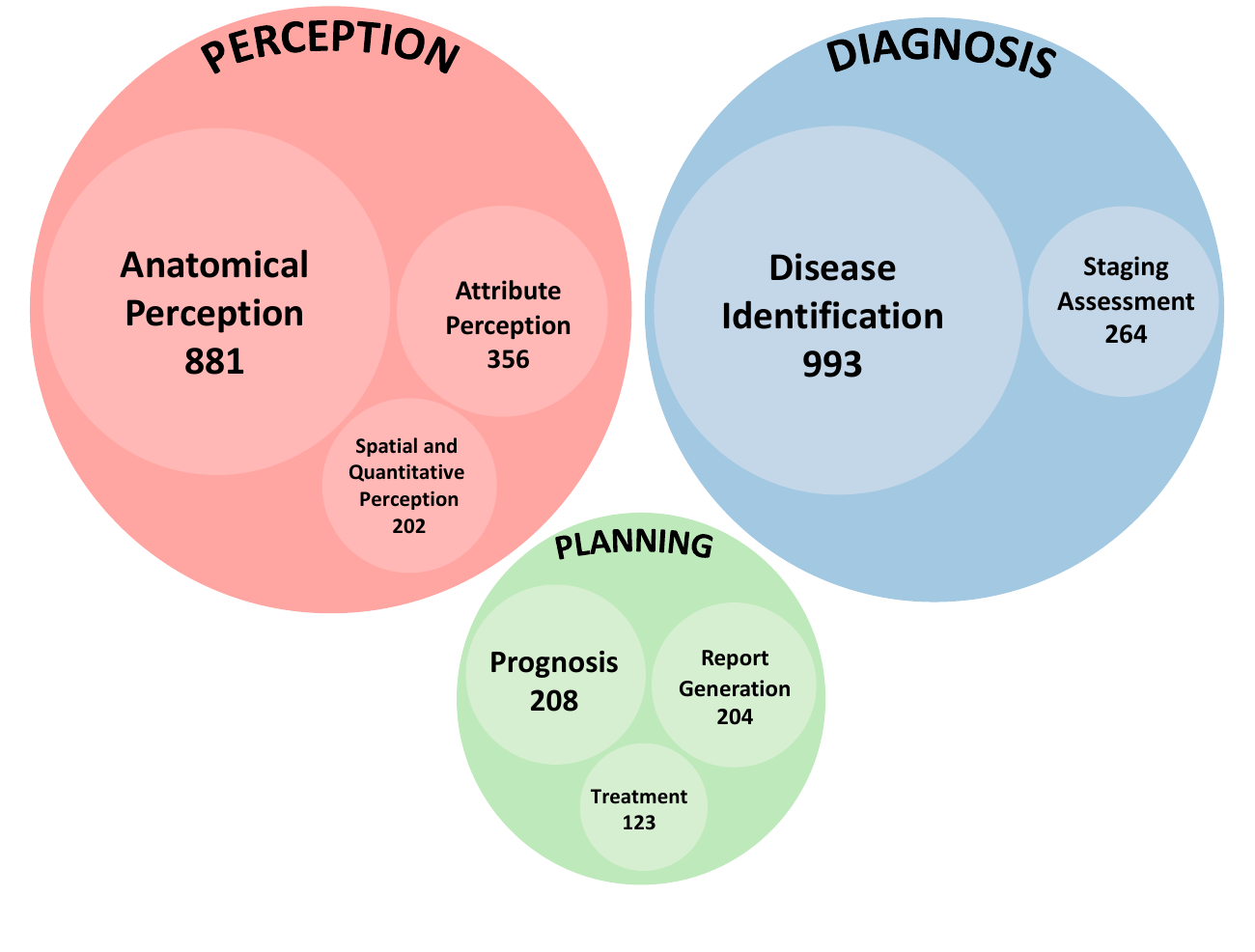}}
 \vspace{-0.35cm}
	\caption{\textbf{Data Statistics for Capacities.} \ourbenchmark~includes two layers of capacity dimensions, which encompass 8 sub-capacities.}
	\label{fig:capacity_sta}
 \vspace{-0.45cm}
\end{figure}

\subsection{Multi-Dimensional Capacity}
\label{sec:capacity}
The decision-making process in clinical practice is multifaceted and layered, encompassing a series of complex cognitive tasks \citep{liberatore2008analytic}. 
The physician typically diagnoses the condition, assesses its stage, and then formulates a treatment plan and prognosis. Subsequently, these decisions and insights should be consolidated into a medical report. To mirror the decision-making intricacies in clinical settings, our benchmark incorporates three primary capacities for Med-MLLMs: \textit{perception}, \textit{diagnosis}, and \textit{planning}. 
Furthermore, we have delineated secondary layer capacities to provide a more granular assessment. From \textit{perception}, we derive (1) \textit{anatomical perception}, (2) \textit{attribute perception}, and (3) \textit{spatial and quantitative perception}. From \textit{diagnosis}, we extract (1) \textit{disease analysis} and (2) \textit{staging assessment}. Lastly, from \textit{planning}, we have identified (1) \textit{treatment}, (2) \textit{prognosis}, and (3) \textit{report generation}.

To illustrate the practical application of these capacities, Figure \ref{fig:specialty} (b) presents various case examples for each capacity within \ourbenchmark. \ourbenchmark~currently encompasses 8 distinct sub-capacities, which offer a multifaceted evaluation of Med-MLLMs' performance in mimicking the decision-making process found in medical practice. Please see Appendix~\ref{sec:append_capacity} for details.
%Supplementary materials provides the detailed definitions of these sub-capacities.

%\begin{figure}[t]
%	\centering
%	\begin{minipage}[t]{0.49\textwidth}
%%		\centering
%\includegraphics[width=\linewidth]{fig/fig_specialty_statistic.pdf}
%	\caption{\textbf{Data Statistics for Specialty}. Currently, \ourbenchmark~incorporates 15 specialties with 3,232 multi-modal questions.}
%	\label{fig:specialty_sta}
%	\end{minipage} 
%	\begin{minipage}[t]{0.49\textwidth}
%%		\centering
%		\includegraphics[width=1.0\linewidth]{fig/fig_capacity_statistic.pdf}
%	\caption{\textbf{Data Statistics for Capacities.} \ourbenchmark~includes two layers of capacity dimensions, which encompass 8 sub-capacities.}
%	\label{fig:capacity_sta}
%	\end{minipage}
%\end{figure}

\subsection{Data Collection}
\label{sec:source_statistics}

To create a comprehensive benchmark with a multifaceted evaluation, our study needs to collect medical images and professional medical knowledge QA pairs, that can effectively test both specialties and capabilities of Med-MLLMs. \ourbenchmark~implements two strategies to generate these QA pairs. %for the generation of
All the data collected in this work has necessary permissions and licenses for research use.
% each aimed at testing different aspects of the Med-MLLMs' proficiency.

The first approach within \ourbenchmark~constructs QA pairs from pre-existing medical image datasets. We collected the test set of 10 different medical vision datasets to cover various question types \citep{liu2022deepdrid,kather_100000_2018,tsuneki_patchgastricadc22_2021,kumar2019multi,yang2023medmnist,shah_breast_2019,acevedo2020dataset,holste2022long,kermany2018identifying,tschandl2018ham10000}. Leveraging the collected datasets, we transformed the original image classification data into a VQA format. To augment the diversity of question types, we constructed several VQA templates for each dataset. For instance, the questions for DeepDRiD \citep{liu2022deepdrid} were formulated as "What is the severity level of diabetic retinopathy in this retinal fundus image?", "Does the retinal fundus image show signs of diabetic retinopathy at Level 3?", and "Select the severity level that best describes the retinal fundus image." Regarding the choice options, we incorporated the true label and three other randomly selected classification labels as the multiple-choice options.
These datasets are restructured into a VQA format, with binary classification tasks formulated as yes/no questions and multi-class classification tasks presented as multiple-choice queries. 

Concurrently, the benchmark employs a second strategy, incorporating QA pairs with images derived from the United States Medical Licensing Examination~\cite{usmle} %\footnote{https://www.amboss.com/us/usmle} 
and current medical textbooks \citep{katzung2004basic,pawlina2018histology,murray2015medical,barrett2010ganong,kumar2014robbins,snell2010clinical,sickles2016acr}. The integration of this content ensures alignment with the rigorous standards required for medical licensure, establishing a high bar for the medical knowledge assessment of Med-MLLMs. We took questions accompanied by images from the USMILE and medical textbook quiz. In instances where the question text contained explanations pertaining to the associated image, we omitted it to ensure that MLLM needs to rely on both textual and visual information to answer the questions accurately. Subsequently, we engaged several medical students to meticulously rewrite and review all the questions, including altering the order of the choices and rephrasing the descriptions of symptoms without affecting the fundamental meaning of the questions. The detailed rewriting and review process is illustrated in Appendix \ref{append:review}. Finally, each question is verified by several senior doctors to ensure quality. This rigorous process ensures that the final questions differ from their original counterparts, further reinforcing the correctness and validity of QA pairs. %question-answer

\noindent\textbf{Statistics.} \ourbenchmark~comprises a total of 3,232 data samples, which span across 15 medical specialties and 8 distinct sub-capacities. The visualizations of these two statistics are exhibited in Figure \ref{fig:specialty_sta} and \ref{fig:capacity_sta}. Moving forward, we will preserve a balanced distribution of questions that fulfill various evaluation dimensions.

\noindent\textbf{Data Split.} To maintain the integrity and the blindness of the evaluation benchmark, we divided \ourbenchmark~into development and test subsets. 100 questions are randomly selected as the development set that is entirely publicly accessible, with ground truth answers for each question. The random selection maintains consistent statistics of development and test sets. Conversely, the test subset is only partially disclosed, with data samples being publicly available without ground truth answers. To ascertain the performance on the test subset, participants are required to submit their predictions to the \ourbenchmark~server for an unbiased assessment.% of Med-MLLMs. 
% for Med-MLLMs
%have partitioned

\begin{figure*}[t]
% {
\centering
	{\includegraphics[width=0.80\linewidth]{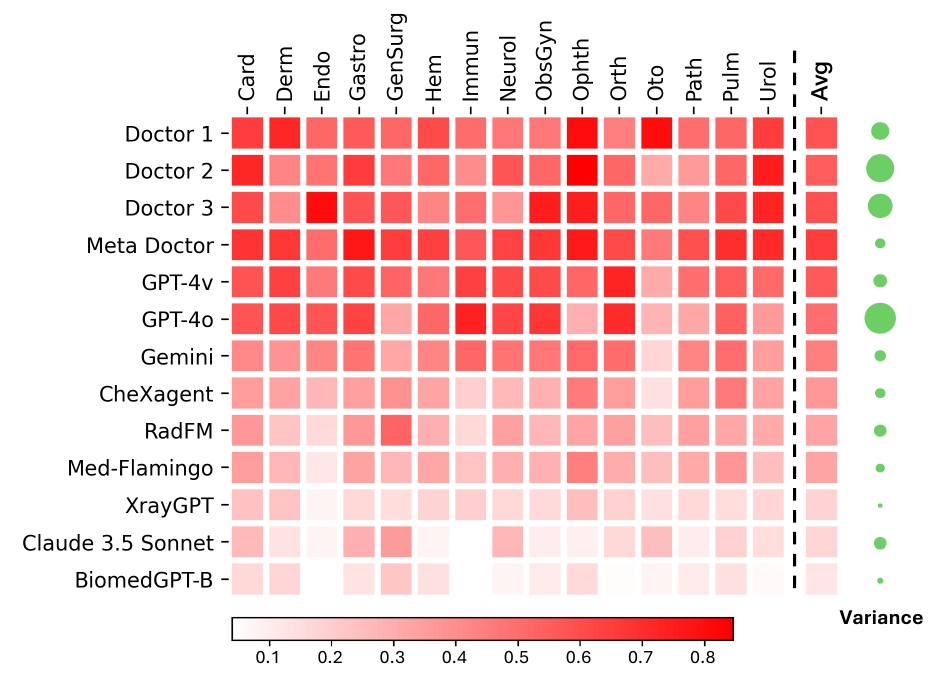}}
 \vspace{-0.2cm}
 % }
 
{\justifying \footnotesize Abbreviation: Card (Cardiology), Derm (Dermatology), Endo (Endocrinology), Gastro (Gastroenterology), GenSurg (General Surgery), Hem (Hematology), Immun (Immunology), Neurol (Neurology), ObsGyn (Obstetrics and Gynecology), Ophth (Ophthalmology), Orth (Orthopedics), Oto (Otolaryngology), Path (Pathology), Pulm (Pulmonology), Urol (Urology).}
% \vspace{-0.1cm}

	\caption{\textbf{The spectrum of Med-MLLMs in Specialties.} Green circle size shows accuracy variance across specialties; larger circles indicate higher variance. Darker squares represent higher accuracy. Numeric details in Appendix Table~\ref{table:models_accuracy}. Meta doctor is ensembled from several doctors whose area of expertise cover these 15 specialties.}% Abbreviation: Card (Cardiology), Derm (Dermatology), Endo (Endocrinology), Gastro (Gastroenterology), GenSurg (General Surgery), Hem (Hematology), Immun (Immunology), Neurol (Neurology), ObsGyn (Obstetrics and Gynecology), Ophth (Ophthalmology), Orth (Orthopedics), Oto (Otolaryngology), Path (Pathology), Pulm (Pulmonology), Urol (Urology).}%\wx{I think the "ave" in Figure 3  could be bold to make it more eye-catching.}}
 
  % The green circle size represents the variance of the accuracy scores across all specialties. The larger circle means a larger variance. The darker color of the square indicates higher accuracy. The detailed digital values refer to supplementary. 
 
	\label{fig:evaluation}
 \vspace{-0.4cm}
\end{figure*}

\section{Experiments}
\subsection{Implementation}
\textbf{Model Evaluation.} This benchmark focuses on four general MLLMs, namely GPT-4V \citep{openai2023gpt4v}, GPT-4o~\citep{openai2024gpt4o}, Gemini \citep{team2023gemini} and Claude 3.5 Sonnet~\citep{claude2024sonnet}, alongside five specialized Med-MLLMs: CheXagent \citep{chen2024chexagent}, RadFM \citep{wu2023generalist}, Med-Flamingo \citep{moor2023medflamingo}, and XrayGPT \citep{thawkar2023xraygpt}, BiomedGPT-B~\citep{zhang2024generalist}. The model information is shown in Appendix Table \ref{tab:llm}. The prompts for these MLLMs are in Appendix Table \ref{tab:prompt}. %, which are recommended in their papers.%, where the prompts for five Med-MLLMs are recommended in their papers.

%\begin{minipage}{\textwidth}
%
%\begin{minipage}[h]{0.52\textwidth}
%\makeatletter\def\@captype{table}
%	\caption{\textbf{The specialty confidence of each doctor.} Use $1 \sim 5$ to represent the confidence score for each specialty. The larger number means more confidence.}
%\scalebox{0.80}{
%\begin{tabular}{l|ccc}
%        \toprule
%        Specialty & Dr. 1 & Dr. 2 & Dr. 3 \\ 
%        \midrule
%        Cardiology & 2 & 5 & 4 \\ 
%        Dermatology & 5 & 2 & 3 \\ 
%        Endocrinology & 3 & 4 & 4 \\ 
%        Gastroenterology & 4 & 4 & 4 \\ 
%        General Surgery & 4 & 3 & 4 \\ 
%        Hematology & 2 & 4 & 3 \\ 
%        Immunology & 5 & 4 & 4 \\ 
%        Neurology & 4 & 4 & 4 \\ 
%        Obstetrics and Gynecology & 2 & 2 & 3 \\ 
%        Ophthalmology & 5 & 2 & 5 \\ 
%        Orthopedics & 4 & 3 & 4 \\ 
%        Otolaryngology & 3 & 2 & 3 \\ 
%        Pathology & 4 & 2 & 3 \\ 
%        Pulmonology & 2 & 4 & 4 \\ 
%        Urology & 4 & 3 & 4 \\ 
%        \bottomrule
%\end{tabular}}
%	\label{table:confidence}
%\end{minipage}
%\begin{minipage}[t]{0.43\textwidth}
%\makeatletter\def\@captype{table}
%\caption{The Modality Study Result. Full information means we use both textual and visual information. w/o means without.}
%\begin{tabular}{>{\arraybackslash}p{3cm}|
%			>{\arraybackslash}p{2cm}}
%\begin{tabular}{>{\arraybackslash}p{2.7cm}|
%			>{\arraybackslash}p{2cm}}
%\begin{tabular}{cc}
%\hline
%Condition & Accuracy \\ \hline
%Full information & 46.2\% \\ 
%w/o image & 32.0\% \\ 
%w/o text & 23.9\% \\ \hline
%\end{tabular}
%
%\label{tab:modality_study}
%\end{minipage}
%\vspace{0.5cm}
%\end{minipage}

\noindent\textbf{Human Study.} To establish a benchmark for performance, the study also includes an evaluation of human specialists. Three clinical specialist doctors are selected to participate, who had expertise in certain medical specialties but were not good at all specialties. Specifically, Doctors 1 through 3 possess 4, 4, and 5 years of professional experience, respectively. Their confidence score for each specialty are listed in Appendix Table \ref{table:confidence}. Moreover, we also invited several doctors whose areas of expertise cover these 15 specialties, and we ensemble their results as the ideal case for meta doctors, who are all-powerful for all specialties.
%The confidence score of these doctors for each specialty is shown in Appendix Table \ref{table:confidence}.
%\wx{I think we can add more details on the doctor, such as the year of clinical and the degree. Also, what is the criteria of being a senior (junior) doctor.}
%A subset of 500 questions is randomly selected from the benchmark. 
% These questions were presented to the clinical doctors, and their responses were recorded. 
%The accuracy of the responses provided by these medical practitioners is calculated to serve as a comparative measure against the performance of the Med-MLLMs.

\noindent\textbf{Evaluation Metrics.} \ourbenchmark~includes a range of question types: multiple choice, yes/no, open-ended questions, and report generation tasks. 
We use accuracy for multiple-choice and yes/no questions and employ GPT to assess open-ended questions. Report generation is evaluated with the ROUGE-L score to measure alignment with gold-standard reports~\citep{wu2023generalist,chen2024chexagent}. Overall accuracy excludes report generation and is calculated as the ratio of correctly answered questions to the total applicable questions in benchmark.
% We adopt accuracy as metric for multiple-choice questions and yes/no questions. Moreover, open-ended questions demand a more subtle assessment approach; here, GPT is utilized to measure the precision of the textual responses. For the evaluation of report generation, the ROUGE-L scoring system is employed to determine the extent to which the models' generated texts align with the gold-standard reports\wt{~\citep{wu2023generalist,chen2024chexagent}}. For the overall accuracy calculation, we consider all questions in the benchmark apart from report generation. The accuracy is computed as the ratio of questions answered correctly to the total number of applicable questions. 
Please see Appendix~\ref{appendix:metrics} for details.

\begin{table}[t]
\renewcommand\arraystretch{1.2}
\caption{\textbf{The Modality Study Result.} Full information means we use both textual and visual information. w/o means without.}
\begin{tabular}{>{\arraybackslash}p{3.5cm}|
			>{\arraybackslash}p{2.5cm}}
\hline
Condition & Accuracy \\ \hline
Full information & 46.2\% \\ 
w/o image & 32.0\% \\ 
w/o text & 23.9\% \\ \hline
\end{tabular}
\label{tab:modality_study}
\end{table}

\subsection{Visual and Textual Modality Study}
To investigate the contribution of each modality, we conducted ablation experiments on GPT-4v. In the first group, we provided GPT-4v with only the visual information and the associated questions. Conversely, in the second group, we supplied GPT-4v with solely the textual information and the questions. The results are shown in Table \ref{tab:modality_study}. The accuracy decline observed for both groups when compared to the baseline condition with access to both modalities proves the importance of both textual and visual information for \ourbenchmark~benchmark. The models cannot answer the questions accurately by solely relying on either the text or visual information alone.

\begin{table*}[h!]
\renewcommand\arraystretch{1.01}
\centering
	\caption{\textbf{The spectrum of Med-MLLMs in Capacity.} Avg* are the average accuracy of Anato, Attr, SpaQua, DisIde, Stag, Prog and Treat. Rep reports the ROUGE-L score. GPT-4V refuses to answer Rep question.} %Abbreviation: Anato (Anatomical Perception), Attr (Attribute Perception), DisIde (Disease Identification), Prog (Prognosis), SpaQua (Spatial and Quantitative Perception), Stag (Staging Assessment), Treat (Treatment Planning), Rep (Report Generation).}
 \vspace{-0.2cm}
\scalebox{0.82}{
	\begin{tabular}{p{0.15\linewidth}|p{0.06\linewidth}p{0.06\linewidth}p{0.08\linewidth}|p{0.06\linewidth}p{0.06\linewidth}|p{0.06\linewidth}p{0.06\linewidth}p{0.06\linewidth}|p{0.06\linewidth}}
		\toprule
		\multirow{2}{*}{Model} & \multicolumn{3}{c|}{Perception} & \multicolumn{2}{c|}{Diagnosis} & \multicolumn{3}{c|}{Planning} & \multirow{2}{*}{Avg*} \\ \cline{2-9} 
		& Anato & Attr & SpaQua & DisIde & Stag & Prog & Treat & Rep & \\ 
		\hline
		% GPT-4V & 0.323  & \textbf{0.385}  & \textbf{0.552}  & 0.649  & \textbf{0.480}  & \textbf{0.504}  & \textbf{0.524}  &N.A.& \textbf{0.462}  \\
		% Gemini & 0.285  & 0.292 & 0.342  & \textbf{0.654}  & 0.342  & 0.496  & 0.323  & 0.082& 0.354  \\
		% CheXagent & 0.238  & 0.253  & 0.321  & 0.524  & 0.252  & 0.451  & 0.315  & \textbf{0.157}& 0.309  \\
		% RadFM & \textbf{0.344}  & 0.298  & 0.212  & 0.130  & 0.396  & 0.295  & 0.290  & 0.091& 0.278  \\
		% Med-Flamingo & 0.270  & 0.256  & 0.217  & 0.587  & 0.272  & 0.398  & 0.145  &0.133& 0.279  \\
		% XrayGPT & 0.163  & 0.107  & 0.152  & 0.082  & 0.104  & 0.223  & 0.145  &0.078& 0.148  \\
  %       Doctor1 \\
  %       Doctor1 \\
  %       Doctor1 \\
  GPT-4V & \textbf{0.462} & \textbf{0.542} & \textbf{0.599} & \textbf{0.592} & 0.504 & 0.649 & 0.556 & N.A & \textbf{0.558} \\ 
GPT-4o & 0.389  & 0.478  & 0.431  & 0.591  & 0.152  & \textbf{0.683}  & \textbf{0.621}  & 0.072  & 0.477 \\
Gemini & 0.410 & 0.410 & 0.460 & 0.386 & \textbf{0.523} & 0.654 & 0.323 & 0.082 & 0.452 \\ 
CheXagent & 0.297 & 0.334 & 0.337 & 0.326 & 0.451 & 0.519 & 0.315 & \textbf{0.157} & 0.368 \\ 
RadFM & 0.378 & 0.404 & 0.455 & 0.219 & 0.295 & 0.082 & 0.290 & 0.091 & 0.303 \\ 
Med-Flamingo & 0.313 & 0.312 & 0.272 & 0.259 & 0.439 & 0.394 & 0.161 & 0.133 & 0.307 \\ 
Claude3.5Sonnet & 0.196  & 0.160  & 0.322  & 0.060  & 0.061  & 0.101  & 0.161  & 0.051  & 0.136 \\
XrayGPT & 0.161 & 0.110 & 0.099 & 0.169 & 0.223 & 0.014 & 0.137 & 0.078 & 0.131 \\ %\hline
% BiomedGPT-S & 0.103  & 0.101  & 0.059  & 0.081  & 0.019  & 0.000  & 0.121  & 0.032  & 0.079 \\
% BiomedGPT-M & 0.089  & 0.101  & 0.045  & 0.078  & 0.068  & 0.010  & 0.153  & 0.019  & 0.079 \\
BiomedGPT-B & 0.089  & 0.110  & 0.040  & 0.104  & 0.098  & 0.010  & 0.129  & 0.071  & 0.090 \\ \hline

Doctor 1 & 0.541 & 0.368 & 0.700 & 0.590 & 0.667 & 0.500 & 0.733 & N.A & 0.586 \\ 
Doctor 2 & 0.523 & 0.435 & 0.467 & 0.575 & 1.000 & 1.000 & 0.556 & N.A & 0.651 \\ 
Doctor 3 & 0.517 & 0.480 & 0.700 & 0.624 & 0.538 & 1.000 & 0.429 & N.A & 0.613 \\ 
Meta Doctor & 0.581 & 0.589 & 0.593 & 0.686 & 0.605 & 1.000 & 0.772 & N.A & 0.689 \\ 
		\bottomrule
	\end{tabular}}
	\label{table:models_accuracy_capacity}
 
	\smallskip
	\footnotesize Abbreviation: Anato (Anatomical Perception), Attr (Attribute Perception), DisIde (Disease Identification), Prog (Prognosis), SpaQua (Spatial and Quantitative Perception), Stag (Staging Assessment), Treat (Treatment Planning), Rep (Report Generation).
 \vspace{-0.6cm}
\end{table*}

\begin{figure*}[t]
\centering
	{\includegraphics[width=0.78\linewidth]{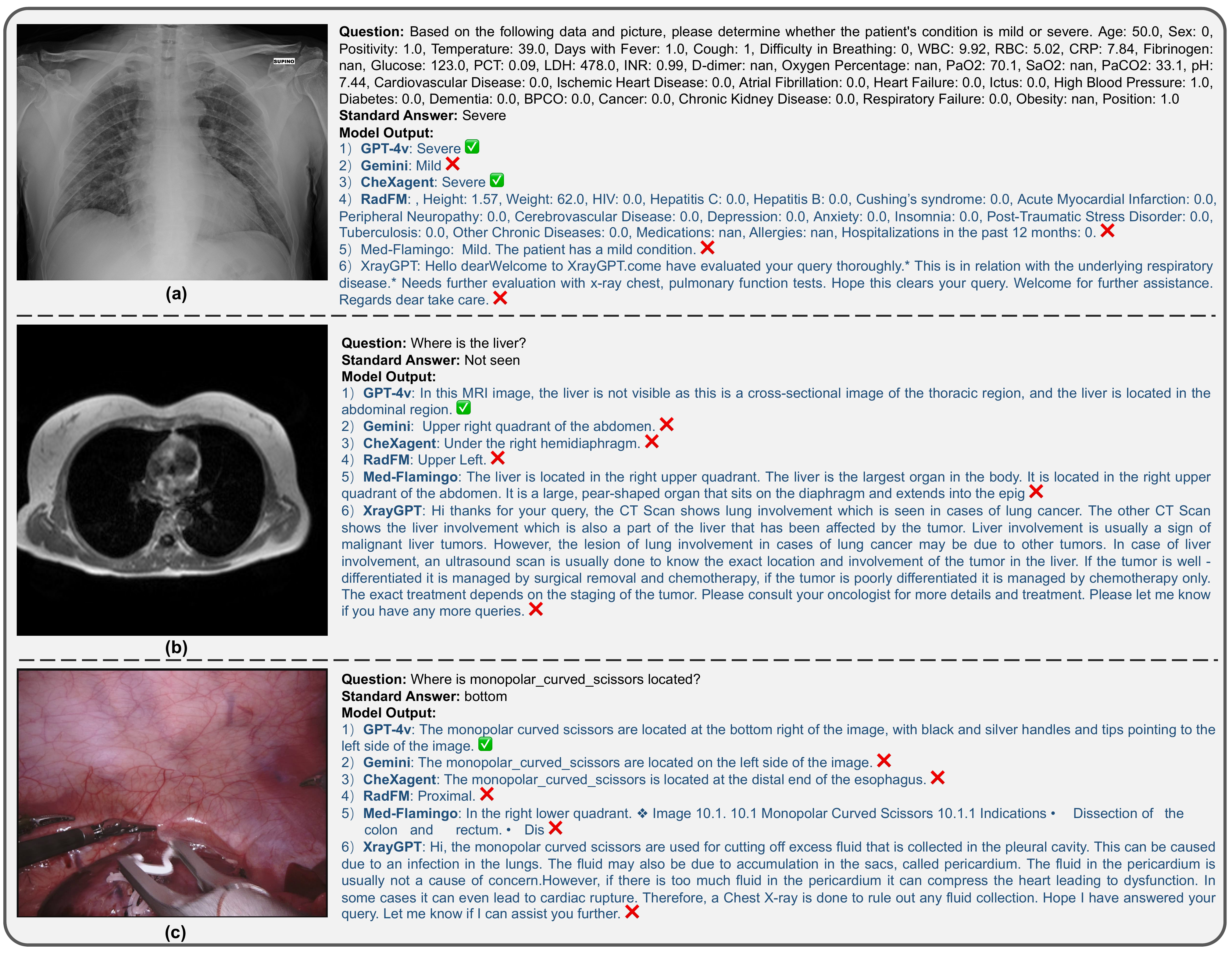}}
 \vspace{-0.1cm}
	\caption{\textbf{Case Study for Common Problem Revealed in Evaluation}. (a) Case for Limited Instruction-following Capabilities. (b) Case for Failed Anatomical Perception. (c) Case for Failed Spatial and Quantitative Perception).}
	\label{fig:instruction}
 \vspace{-0.45cm}
\end{figure*}

\subsection{Results across Specialties}

\label{sec:three_insight}

Our analysis reveals a broad spectrum of accuracy scores across the medical specialties in Figure \ref{fig:evaluation}. GPT-4V generally has the highest performance among the Med-MLLMs, with its average accuracy closest to human doctors. GPT-4o has the second-best performance among the Med-MLLMs, with moderate accuracy, but still significantly lower than GPT-4V.
% Gemini has the second-best performance among the Med-MLLMs, with moderate accuracy, but still significantly lower than GPT-4V. 
CheXagent, RadFM, Med-Flamingo, XrayGPT, and BiomedGPT show lower performance across all specialties. When comparing between Med-MLLMs and human doctors, human doctors generally outperform Med-MLLMs in all specialties. However, Med-MLLMs demonstrate a markedly lower variability in their performance across specialties when compared to humans.
%However, Med-MLLMs demonstrate a markedly lower variability in their performance across medical specialties when compared to humans. 
%it is evident that 

\subsection{Results across Capacities}

The spectrum of Med-MLLMs in capacity is shown in Table \ref{table:models_accuracy_capacity}. GPT-4V leads with the highest average accuracy across the seven tasks with an average score of 55.8\%. This indicates that GPT-4V is likely the most versatile and reliable model for a range of medical tasks. 
% Notably, some Med-MLLMs perform better than GPT-4V in certain capacities. For example, RadFM has the highest score in Anatomical Perception with 34.4\%, indicating that it might be the best at identifying anatomical structures. 
Moreover, both GPT-4V and GPT-4o perform equally well in Disease Identification with a score of 59.2\% and 59.1\%, respectively, indicating strong capabilities in diagnosing diseases.
Notably, GPT-4o achieves better performance than GPT-4V in prognosis and treatment planning with accraucy of 68.3\% and 62.1\%, suggesting its significant ability in planning tasks.
As for the report generation task, GPT-4V refused to generate the responses (see Appendix~\ref{appendix:eval_models} for details)
%\footnote{GPT-4V generated `I'm sorry, but I cannot provide the service of generating radiology reports. If you have any other questions or need information on a different topic, feel free to ask.'}
, which could be due to the model's guidelines or limitations in this specific task. CheXagent has the highest ROUGE-L score for Report Generation at 0.157, although this is still relatively low, suggesting room for improvement in how these models generate medical reports. %\wx{Reviewer may ask why chatgpt only refuses to answer report generation tasks.}

\section{Discussion}
\label{sec:suggestion}
From the above results, six key insights have been deduced as follows:

\noindent\textbf{1) Significant variance exists in different specialties.} Human doctors have strong performances in certain specialties, but weaker performances in others. As illustrated in Appendix Table \ref{table:confidence}, the confidence score assigned by Doctor2 is 5 for Cardiology compared to a score of 2 for Dermatology. Correspondingly, the results indicates that Doctor2 achieved a diagnostic accuracy of 71.4\% for Cardiology, which markedly contrasts with a reduced accuracy of 40.0\% for Dermatology. These variations highlight the complexity and diversity inherent to each medical specialty. Given these differences, it is critical to establish a comprehensive Med-MLLMs benchmark to systematically evaluate the performance across various specialties.

\noindent\textbf{2) Human doctors outperform Med-MLLMs.} Human physicians surpass Med-MLLMs in diagnostic accuracy across all specialties. Even a junior doctor with an average accuracy of 57.4\%, marginally exceeds the most proficient Med-MLLM, GPT-4V, which achieved an accuracy of 54.3\%. This outcome suggests that, despite the advancements made in artificial intelligence, there is still a gap in diagnostic precision between human expertise and current Med-MLLMs. This underscores the need for continued development and specialization in the field of AI-driven medical diagnostics. On the other hand, the GPT-4V is comparable with junior human doctors, which indicates that Med-MLLMs like GPT-4V have the potential to complement the diagnostic process in practice. However, the current results indicate that the integration of such models into clinical workflows should be approached with caution, ensuring they serve as an adjunct to, rather than a replacement for, human clinical judgement.

\noindent\textbf{3) Superiority of Generalist MLLMs Over Specialized Med-MLLMs.} Results reveal that generalist models such as GPT-4V and Gemini outperform five specialized Med-MLLMs in a dual-spectrum evaluation. Notably, RadFM, despite being trained on 16 million multi-modal medical question-answer pairs, remains inferior to GPT-4V. According to Appendix Table \ref{tab:llm}, the parameter count of the general MLLMs are much larger than that of Med-MLLMs. Therefore, we recommend that future research on Med-MLLMs should explore increasing the model parameter capacity to potentially improve their performance.
%\wx{I think the difference between gpt4 and specialized Med-MLLMs are not only general v.s. Specialized. This highlight may lead to some concern from reviewers. Can we just say the Superiority of recent MLLMs}

\noindent\textbf{4) Limited Long-range Instruction Capture.} Despite the implementation of careful prompt engineering, certain Med-MLLMs exhibit a tendency to generate indirect responses. Figure \ref{fig:instruction}(a) illustrates instances where, instead of answering the condition as prompted, models like RadFM and XrayGPT provide irrelevant information. We argue that an optimal Med-MLLM should adhere to such detailed instructions.

\noindent\textbf{5) Limited Multi-Modality Fusion.} Figure \ref{fig:instruction} (b) illustrates the case that only GPT-4V accurately incorporates image information into its response. In contrast, other Med-MLLMs simply restate the well-known fact that the liver is located in the upper right quadrant of the abdomen, neglecting to integrate the visual data presented. This pattern suggests a limited ability of most Med-MLLMs to synthesize image and text information, as they solely rely on textual prompts for their answers. Enhancing Med-MLLMs' multi-modal fusion capabilities emerges as a crucial and promising direction for future development.

\noindent\textbf{6) Med-MLLMs offer the potential for integration of expansive and in-depth medical knowledge.} In contrast to the wider variability observed among human physicians, Med-MLLMs exhibit more uniform performance, as depicted by the smaller green circles in Figure \ref{fig:evaluation}. This reduced variance signifies a standardized diagnostic capability of Med-MLLMs across different medical specialties. Such a pattern suggests that while human specialists are in-depth for particular domains, Med-MLLMs provide a more expansive knowledge across diverse medical fields, which could potentially be leveraged to augment clinical decision-making processes. This is particularly relevant for complex multi-system disorders like hypermobile Ehlers-Danlos syndrome (hEDS), where an interdisciplinary approach is paramount \citep{gensemer2021hypermobile}. The advent of Med-MLLMs makes the integration of expansive and in-depth medical knowledge feasible, offering the potential to address the multifaceted needs of patients.

\section{Conclusion}

We introduce~\ourbenchmark, a comprehensive Med-MLLM evaluation benchmark with 3,232 multi-modal questions spanning 15 medical specialties and 79 body parts/organs, for specialty and capacity analysis. It includes a website for secure server-side evaluation of submitted results. We assess 6 Med-MLLMs and 5 human doctors on~\ourbenchmark. Analysis shows that while current Med-MLLMs have limitations, they can supplement clinical judgment, suggesting potential for integrated medical knowledge application across breadth and depth.
%, to analyze the spectrum in specialty and capacity
% for specialty and capacity analysis

% In this paper, we introduce a comprehensive evaluation benchmark for Med-MLLMs, \ourbenchmark, comprising 3,232 multi-modal questions, encompassing 15 medical specialties and 79 distinct body parts and organs, to analyze the spectrum in specialty and capacity. Additionally, we have developed a website that allows for the submission and server-side evaluation of predictive results to ensure integrity. 6 Med-MLLMs and 5 human doctors are evaluated in \ourbenchmark. After analyzing the performance, we find that while the current Med-MLLMs have limitations, they possess the capability to supplement human clinical judgment, suggesting the possibility for integrated medical knowledge application that spans both breadth and depth.

\section{Limitations}
This paper has two primary limitations that offer avenues for future research:
\begin{itemize}[leftmargin=*]
\item The current benchmark does not consider long patient history narratives that are often crucial for real-world clinical decision-making. As future work, we plan to expand the question set to incorporate queries that require comprehending and reasoning over long sequences of patient records. This will allow for a more comprehensive evaluation of Med-MLLMs' performance in scenarios that better approximate the complexities of actual clinical practice involving longitudinal patient data.
\item The questions of Perception/Diagnosis/Planning are independent of each other currently, without coherence. Real clinical decision-making needs to be completed coherently from the front end to the back end. If one of them is wrong, the diagnosis will be not correct. In the future, the question sets will integrate long sequences of data in patient records and provide sequential disease questions for a patient.
\end{itemize}

\bibliography{custom}

\clearpage
\appendix

\noindent { \LARGE \textbf{Appendix for \ourbenchmark}}

\smallskip
\smallskip
\smallskip

\noindent\textbf{Abstract.} 
In this supplementary material, we provide additional information about the \ourbenchmark.
%Appendix~\ref{sec:novelty_analysis} offers the novelty analysis of the proposed CLIP-Driven Universal Model. 
Appendix~\ref{sec:append_specialty} illustrates the definition of each specialty. 
Appendix~\ref{append:review} provides the details of re-writing and review process.
Appendix~\ref{sec:organ} provides list of involved organ. Appendix~\ref{sec:append_capacity} elaborates on the capacity taxonomy.
In Appendix~\ref{sec:example}, we provide some examples in \ourbenchmark~and some case studies of Med-MLLMs for each capacity.
Finally, Appendix~\ref{sec:qqr} supplements the qualitative and quantitative results in the main paper, including the visualization of statistics for specialty and capacity, and the digital results of different specialties.

\section{Specialty List}
\label{sec:append_specialty}

\ourbenchmark~encompasses 15 medical specialties that represent the core divisions of modern healthcare. These specialties can be broadly categorized into several groups based on their focus areas and interconnected nature. They are Hematology, Cardiology, Endocrinology, Obstetrics and Gynecology, Gastroenterology, Immunology, Urology, Orthopedics, Neurology, Otolaryngology, Pulmonology, Dermatology, Pathology, Ophthalmology, General surgery.

\noindent\textbf{Internal Medicine Specialties:}
\textit{Cardiology} focuses on the heart and cardiovascular system, diagnosing and treating various heart conditions. \textit{Pulmonology} addresses disorders of the respiratory system, including the lungs and airways. \textit{Gastroenterology} specializes in the digestive system, managing disorders from the esophagus to the intestines. \textit{Endocrinology} deals with the endocrine system and hormone-related disorders, while \textit{Hematology} focuses on blood, blood-forming organs, and related diseases. \textit{Immunology} studies the immune system and its disorders, playing a crucial role in understanding the body's defense mechanisms.

\noindent\textbf{Surgical and Procedural Specialties:}
\textit{General Surgery} encompasses procedures involving abdominal organs, endocrine glands, and various soft tissues. \textit{Orthopedics} concentrates on the musculoskeletal system, treating bone, joint, and muscle disorders. \textit{Urology} addresses both the urinary tract and male reproductive system. \textit{Otolaryngology} (ENT) specializes in ear, nose, and throat conditions, while \textit{Ophthalmology} focuses exclusively on eye diseases and vision care.

\noindent\textbf{Specialized Care and Diagnostics:}
\textit{Obstetrics and Gynecology} provides comprehensive care for women's reproductive health and pregnancy. \textit{Dermatology} focuses on conditions affecting the skin, hair, and nails. \textit{Neurology} addresses disorders of the nervous system, including the brain and spinal cord. \textit{Pathology}, serving as a diagnostic cornerstone, studies disease processes through tissue and fluid examination.

\section{Organ list}
\label{sec:organ}

This section we list the involved body parts and organs in this benchmark.

53 Body Parts: Abdomen, Neck, Chest, Head, Cervical Vertebrae, Ankle, Femur, Vertebrae, Bowel, Mandible, Knee, Cochlea, Hand, Bladder, Spine, Wrist, Pelvis, Carotid Artery, Carotid Bifurcation, Trachea, Larynx, Colorectal, Blood, Forearm, Elbow, Hip, Muscle, Reproductive Systems, Gastrointestinal Tract, Ligamentum Nuchae, Small Intestine, Colon, Seminal Vesicle, Duodenum, Anterior Pituitary, Parathyroid, Vena-cava, Right Atrium, Left Ventricle, Muscular Artery, Bronchiole, Aorta, Palatine Tonsil, Pyloric Stomach, Cardiovascular, Endocrine, Musculo skeletal, Ophthalmic, Pulmonary, Blood smear, Cartilage, Adipose Tissue, Tendon,  Nervous Tissue, 

26 Organs: Liver, Lung, Brain, Breast, Testicle, Thyroid, Ovaries, Kidney, Heart, Uterus, Intestine, Pancreas, Pituitary gland, Stomach, Gallbladder, Skin, Eye, Blood, Ileum, Lymph Node, Umbilical Cord, Prostate, Duodenum, Parathyroid, Esophagus, Appendix

\section{Capacity}
\label{sec:append_capacity}

\subsection{Perception}

\subsubsection{Anatomical Perception}
Anatomical Perception is the ability to recognize and understand the normal structures of the body, including their locations, sizes, shapes, and the planes in which they are imaged.

Anatomical Perception is foundational to medical imaging analysis, as it allows for the accurate identification of body parts and serves as a basis for detecting abnormalities. It involves discerning the detailed anatomy within complex images and is essential for any subsequent diagnostic or therapeutic action.

\subsubsection{Attribute Perception}
Attribute Perception is the capacity to discern various attributes of tissues and structures, such as their density, texture, composition, and presence of pathological signs, along with recognizing instruments, modalities, and colors when applicable.

Attribute Perception focuses on the finer details that characterize tissues and help differentiate between normal and abnormal findings. It is vital for quality imaging interpretation and aids in the detailed understanding of pathologies and their implications on health.

\subsubsection{Spatial and Quantitative Perception}
Spatial and Quantitative Perception encompasses the skill to evaluate the spatial relations and quantitative aspects within medical images, such as counting entities and understanding their three-dimensional positions and relationships.

Spatial and Quantitative Perception is crucial for tasks that require an understanding of the geometry and distribution of anatomical structures and pathological findings, which is important for accurate diagnosis, surgical planning, and treatment evaluation.

\subsection{Diagnosis}

\subsubsection{Disease Analysis}
Disease Analysis is the ability to accurately detect and diagnose diseases from medical data, which may include imaging, laboratory results, patient history, and physical examination findings.

Disease Identification is the cornerstone of clinical practice, where the model's prowess in recognizing patterns and correlating them with potential diseases is essential. This involves not only identifying the presence of a disease but also understanding its nature.

\subsubsection{Staging Assessment}

Staging Assessment is the process of determining the extent or severity of a disease, especially cancer, within the body, which is crucial for choosing the most appropriate treatment strategy.

Staging Assessment evaluates how far a disease has progressed and is a critical step in the treatment planning process. It requires a comprehensive analysis of data to accurately classify the stage, which can significantly affect prognosis and treatment choices.

\subsection{Planning}

\subsubsection{Treatment}

Treatment is the selection and administration of the appropriate therapeutic interventions tailored to the individual patient's disease and condition.

The Treatment aspect involves creating a detailed plan for managing the patient's condition, which may include medication, surgery, lifestyle changes, or other therapies. The objective is to select the most effective and least invasive treatment options while considering the patient's preferences and overall health status.

\subsubsection{Prognosis}

Prognosis is the prediction of the likely course and outcome of a disease, taking into account the nature of the disease, the patient's general physical condition, and the treatment options available.

Prognosis is about looking ahead to predict possible outcomes for the patient. This includes estimating survival rates, potential complications, and the likelihood of disease recurrence. It is a vital part of patient counseling and informs decision-making for both clinicians and patients.

\subsubsection{Report generation}

Report generation is the synthesis of medical data and analytical findings into coherent, standardized, and actionable reports for use by healthcare providers.

Report Generation combines all the collected information into a format that is understandable and useful for guiding clinical decisions. It ensures that the insights gained from the model's analysis are communicated effectively, serving as a bridge between the model's output and clinical action steps.

\section{Review Process}
\label{append:review}
For the rewriting phase, we recruited 31 medical students, each handling approximately 104 original question-answer pairs. These students were required to meticulously rewrite both the question and answer components following strict revision criteria. When rewriting questions, they had to (1) preserve the fundamental medical meaning, (2) modify at least 10\% of the content, and (3) maintain clarity and professional language. When rewriting answers, they were required to (1) maintain medical accuracy, (2) preserve the original meaning while rephrasing the text, and (3) ensure consistency with the modified question, though they were permitted to reorder the answer choices. In the review phase, we recruited three senior doctors who independently evaluated each pair for medical accuracy and consistency.

Three senior doctors were engaged in the review phase to ensure medical accuracy and consistency. For each question, reviewers were provided with both the original and rewritten versions, and doctors independently evaluated two aspects: question consistency (whether the rewritten question maintains the same medical intent as the original) and answer correctness (whether the provided answer is medically accurate). If two or more doctors identified inconsistency or inaccuracy (in either consistency or correctness evaluation), the question underwent a collaborative review and revision process involving all three senior doctors. This consensus-based approach ensured high-quality question-answer pairs. Out of 3,232 questions in our benchmark, 127 questions underwent revision based on doctor feedback, and the inter-rater reliability coefficient among the three doctors was 0.78, indicating substantial agreement.

\section{Examples in \ourbenchmark}
\label{sec:example}
\begin{figure*}[t]
	{\includegraphics[width=\textwidth]{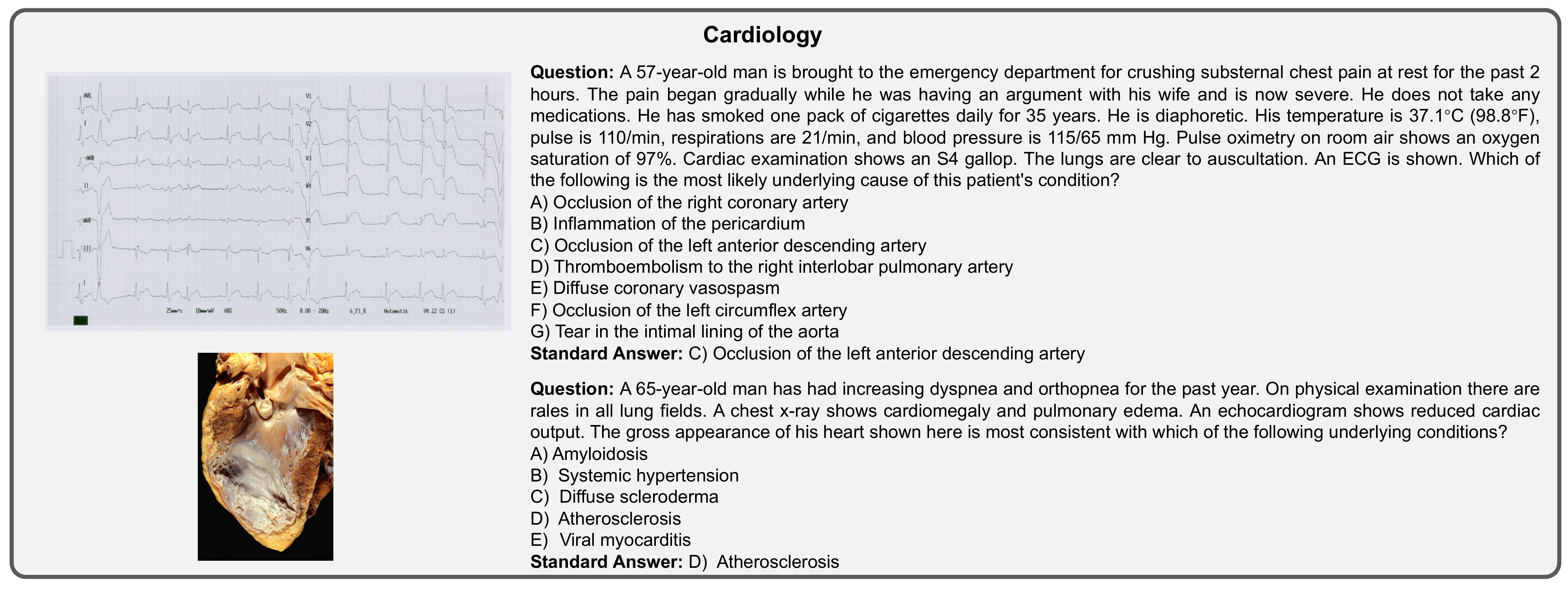}}
	\caption{\textbf{Examples for Cardiology.}}
	\label{fig:capacitycase1}
\end{figure*}

\begin{figure*}[t]
	{\includegraphics[width=\textwidth]{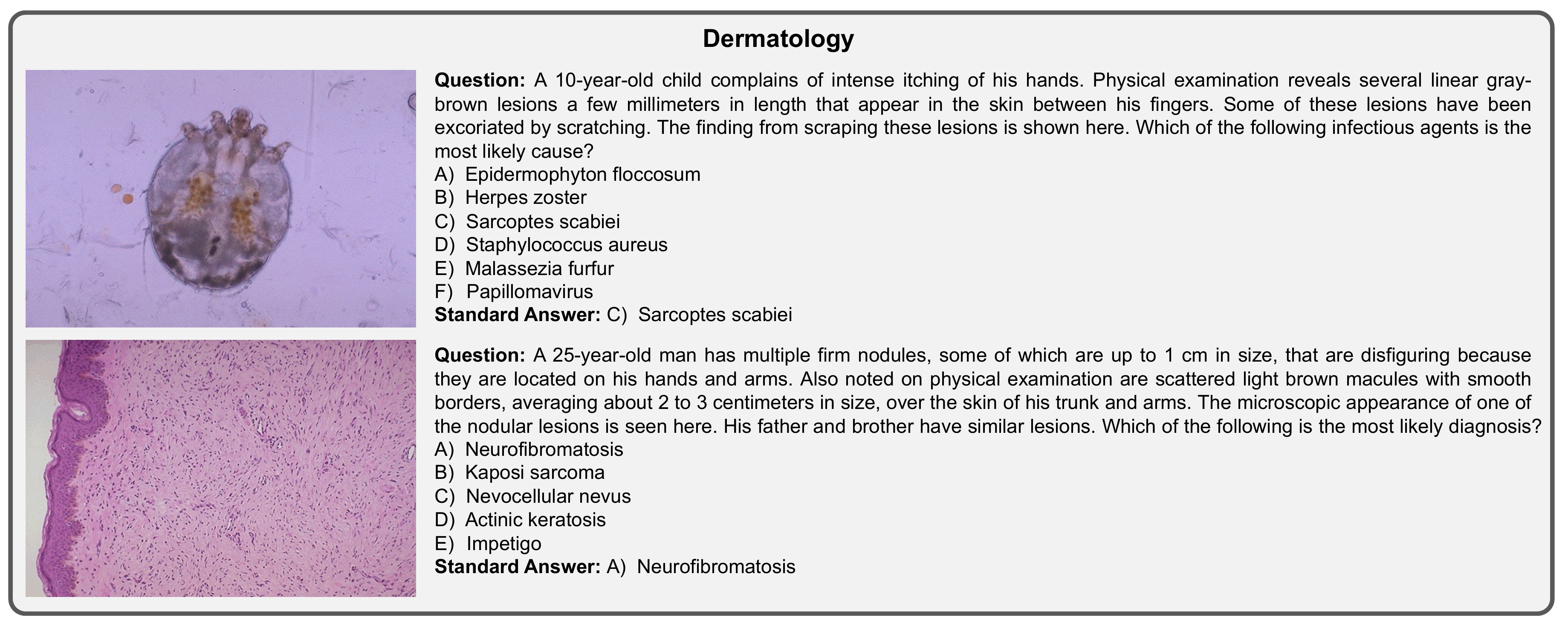}}
	\caption{\textbf{Examples for Dermatology.}}
	\label{fig:capacitycase2}
\end{figure*}

\begin{figure*}[t]
	{\includegraphics[width=\textwidth]{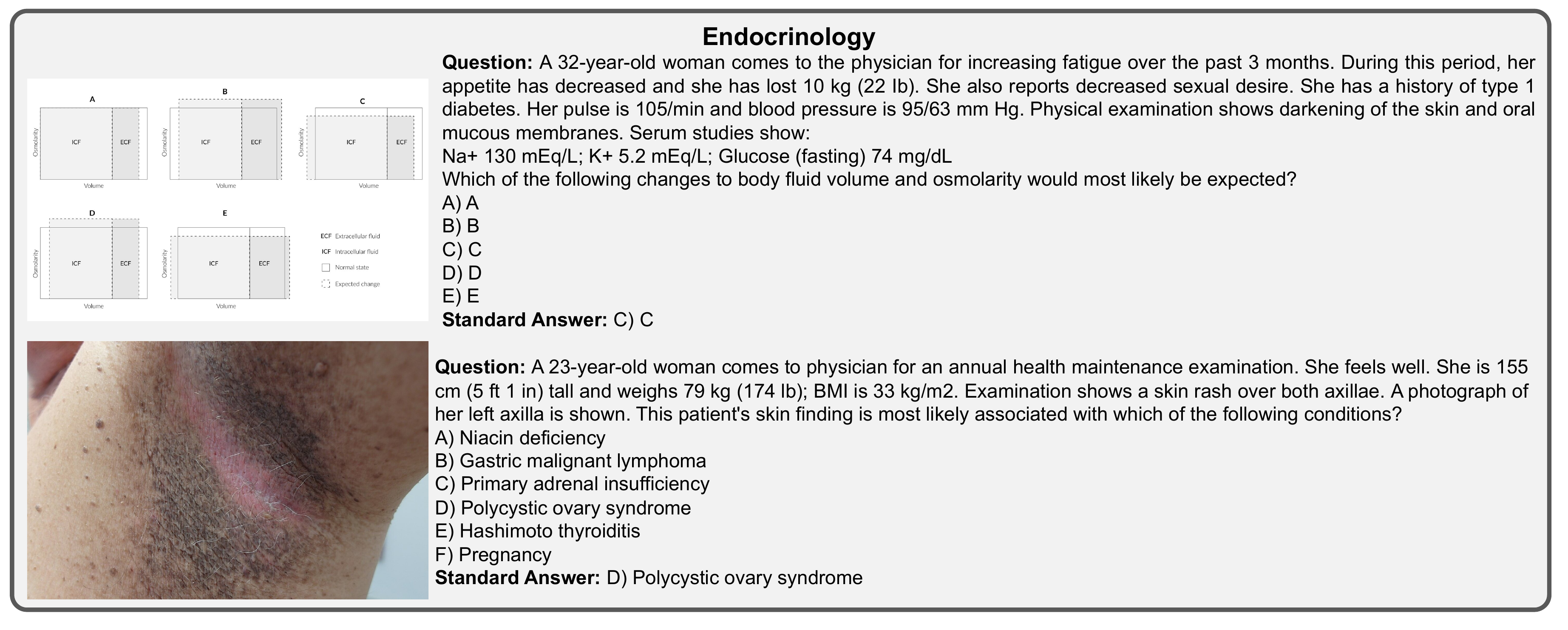}}
	\caption{\textbf{Examples for Endocrinology.}}
	\label{fig:capacitycase3}
\end{figure*}

\begin{figure*}[t]
	{\includegraphics[width=\textwidth]{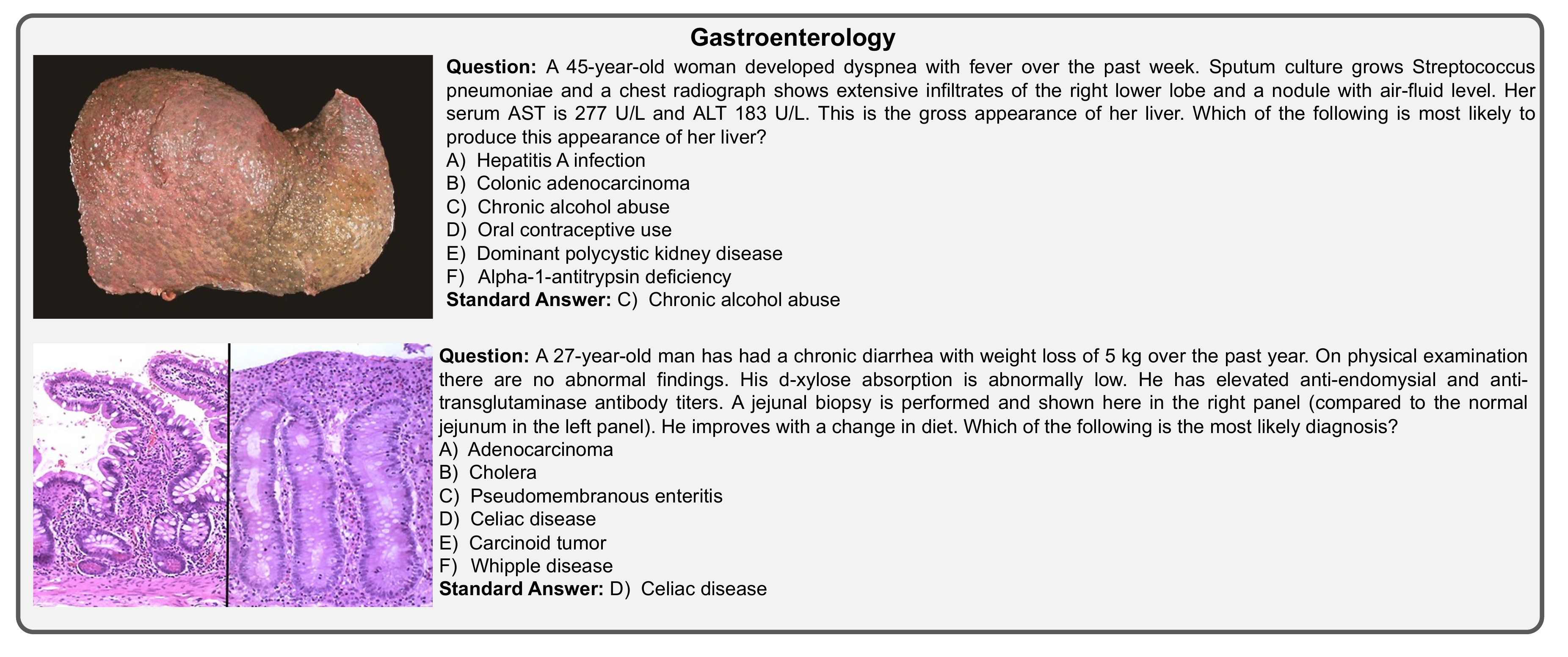}}
	\caption{\textbf{Examples for Gastroenterology.}}
	\label{fig:capacitycase4}
\end{figure*}

\begin{figure*}[t]
	{\includegraphics[width=\textwidth]{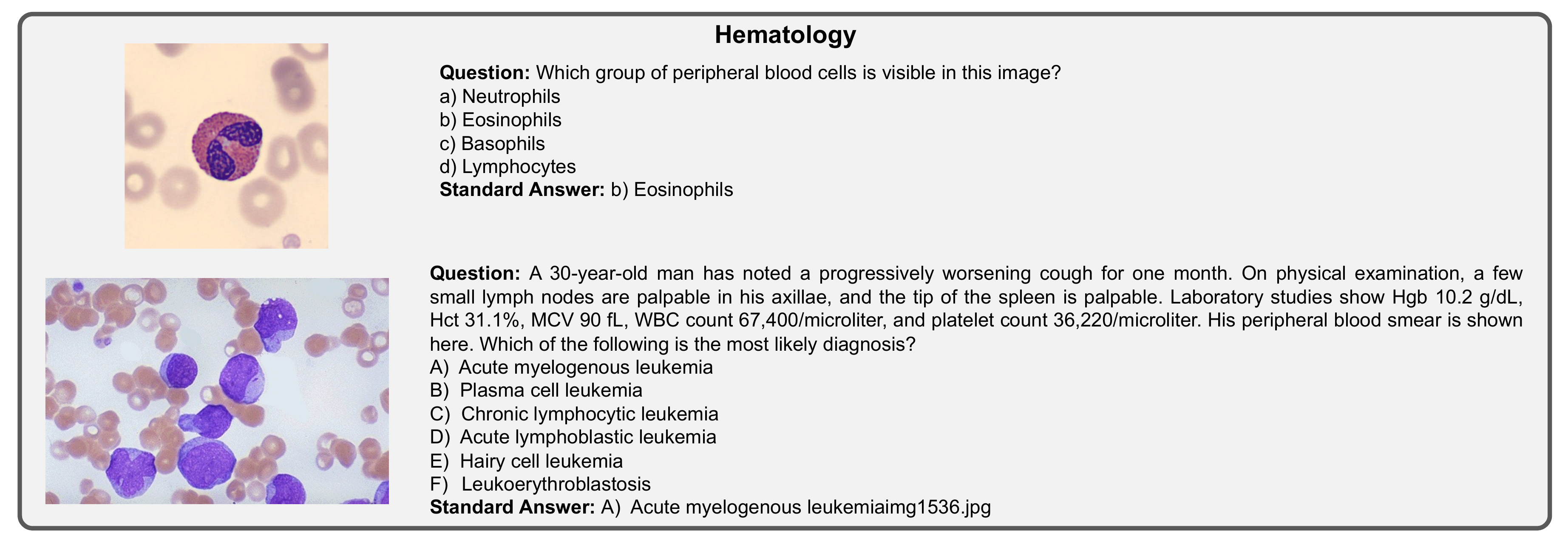}}
	\caption{\textbf{Examples for Hematology.}}
	\label{fig:capacitycase5}
\end{figure*}

\begin{figure*}[t]
\centering
	{\includegraphics[width=1.0\linewidth]{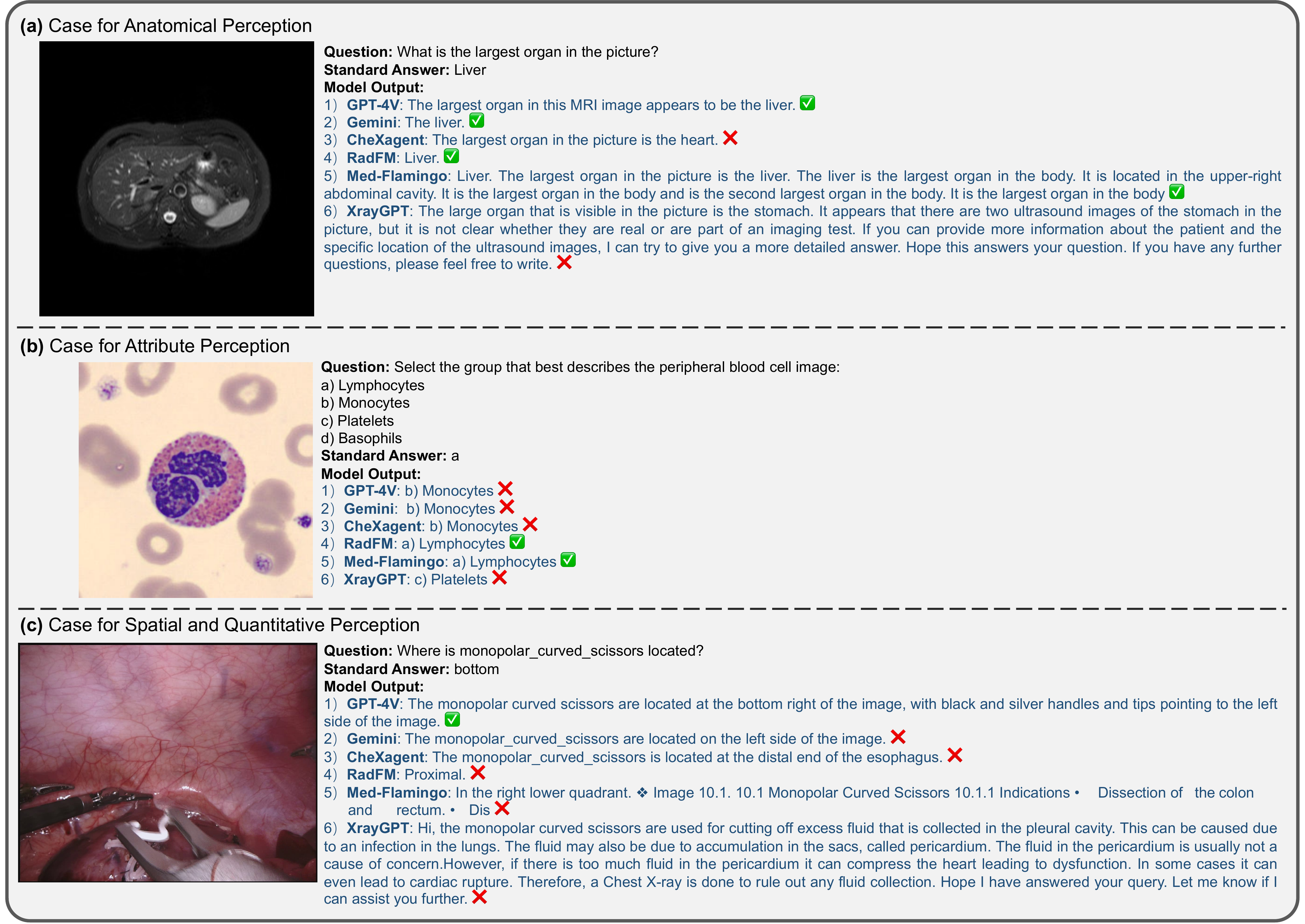}}
	\caption{\textbf{Case study for perception.}}
	\label{fig:capacity_percep}
\end{figure*}

\begin{figure*}[t]
\centering
	{\includegraphics[width=1.0\linewidth]{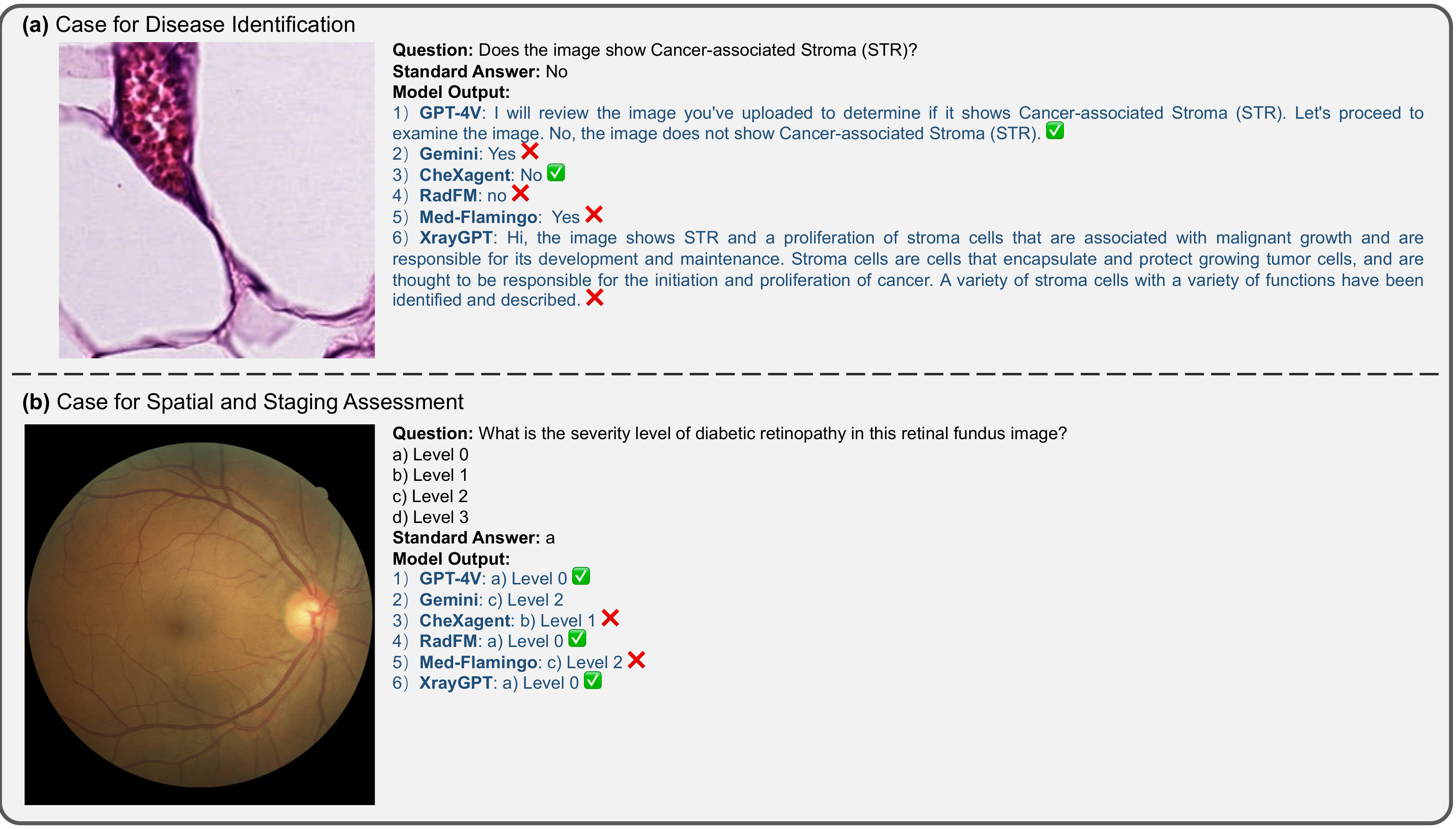}}
	\caption{\textbf{Case study for diagnosis.}}
	\label{fig:capacity_diagno}
\end{figure*}

\begin{figure*}[t]
\centering
	{\includegraphics[width=1.0\linewidth]{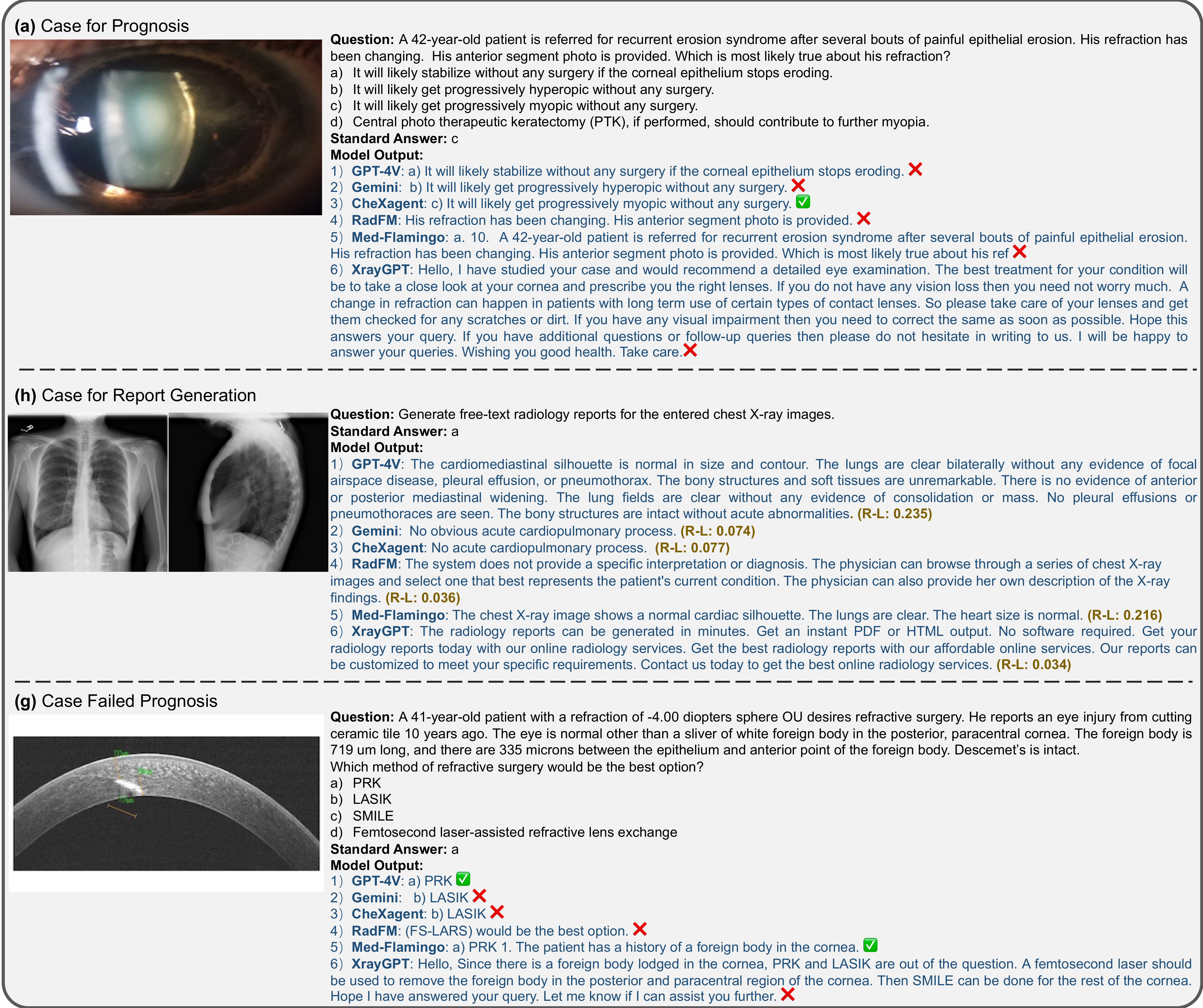}}
	\caption{\textbf{Case study for planning.} R-L represents the ROUGE-L score.}
	\label{fig:capacity_plan}
\end{figure*}

In Figures \ref{fig:capacitycase1} to \ref{fig:capacitycase5}, we illustrate several examples in different specialties in \ourbenchmark. Also, we provide some case studies for various capacities in Figures \ref{fig:capacity_percep} to \ref{fig:capacity_plan}.

\begin{table}[t]
\renewcommand\arraystretch{1.2}
	\caption{\textbf{The specialty confidence of each doctor.} Use $1 \sim 5$ to represent the confidence score for each specialty. The larger number means more confidence.}
 \scalebox{0.95}{
	\begin{tabular}{l|ccc}
        \toprule
        Specialty & Dr. 1 & Dr. 2 & Dr. 3 \\ 
        \midrule
        Cardiology & 2 & 5 & 4 \\ 
        Dermatology & 5 & 2 & 3 \\ 
        Endocrinology & 3 & 4 & 4 \\ 
        Gastroenterology & 4 & 4 & 4 \\ 
        General Surgery & 4 & 3 & 4 \\ 
        Hematology & 2 & 4 & 3 \\ 
        Immunology & 5 & 4 & 4 \\ 
        Neurology & 4 & 4 & 4 \\ 
        Obstetrics and Gynecology & 2 & 2 & 3 \\ 
        Ophthalmology & 5 & 2 & 5 \\ 
        Orthopedics & 4 & 3 & 4 \\ 
        Otolaryngology & 3 & 2 & 3 \\ 
        Pathology & 4 & 2 & 3 \\ 
        Pulmonology & 2 & 4 & 4 \\ 
        Urology & 4 & 3 & 4 \\ 
        \bottomrule
\end{tabular}}
	\label{table:confidence}
\end{table}

\section{Qualitative and Quantitative Results}
\label{sec:qqr}
This section supplements the results of evaluation: 

Table \ref{table:confidence} shows the confidence of doctors in various specialties.

We performed ablation experiments on GPT-4v to assess each modality's contribution. One group receives only visual information and questions, while another receives only text and questions. As shown in Appendix Table \ref{tab:modality_study}, both groups' accuracy decline compared to the baseline with both modalities, highlighting the necessity of combining textual and visual information for the \ourbenchmark~benchmark. Models cannot accurately answer questions using only text or visual data alone.

As shown in Table \ref{table:models_accuracy}, the human doctors' performance varies widely across specialties and individuals, with accuracy scores ranging from 0.286 to 0.846 in the different fields. The average accuracy of human doctors (calculated from the "Avg" column) ranges from 0.538 to 0.641, with Meta-Doctor having the highest average accuracy. Comparing the Med-MLLMs to human doctors, all the Med-MLLMs have a lower average accuracy than the human doctors. Among the Med-MLLMs, GPT-4V's performance is closest to that of human doctors, which surpasses Doctor2 by a small margin with an accuracy of 0.005. 

\begin{figure*}[t]
	{\includegraphics[width=\textwidth]{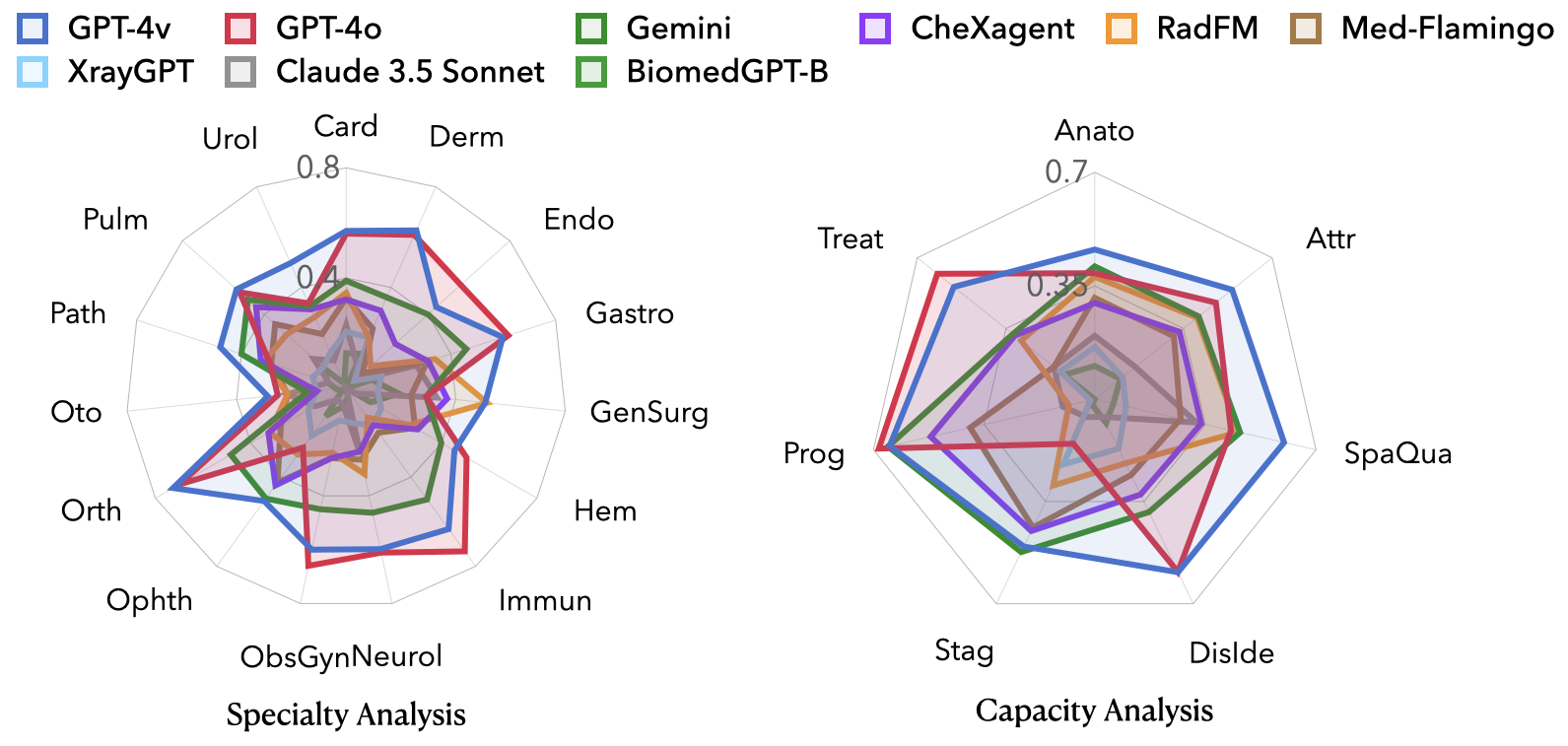}}
	\caption{\textbf{Comparison of Med-MLLMs on our benchmark.}}
	\label{fig:radar}
\end{figure*}

Figure \ref{fig:radar} visualizes the performance comparison of Med-MLLMs on the benchmark. Each model has its strengths and weaknesses, with GPT-4V showing the most robust performance across most areas. Gemini seems to be a decent second choice, particularly in prognosis. The other models have niche areas where they perform well but are generally outperformed by the top two models.

\section{Evaluation Models}
\label{appendix:eval_models}
% \ourbenchmark~focuses on two general MLLMs, including GPT-4V \citep{openai2023gpt4v} and Gemini \citep{team2023gemini}, along with four specialized Med-MLLMs, i.e. CheXagent \citep{chen2024chexagent}, RadFM \citep{wu2023generalist}, Med-Flamingo \citep{moor2023medflamingo}, and XrayGPT \citep{thawkar2023xraygpt}. We evaluate the GPT-4V and Gemini through the official API. We test the rest of the specialized Med-MLLMs through the released code and pre-trained model.
\ourbenchmark~focuses on four general MLLMs, including GPT-4V, GPT-4o, Gemini and Claude 3.5 Sonnet, along with five specialized Med-MLLMs, i.e. CheXagent, RadFM, Med-Flamingo, XrayGPT, and BiomedGPT, as shown in Table~\ref{tab:llm}. We evaluate the GPT-4V, GPT-4o, Gemini and Claude 3.5 Sonnet through the official API. We test the rest of the specialized Med-MLLMs through the released code and pre-trained model. The prompt for these MLLMs are listed in Table~\ref{tab:prompt}, where the prompts for five Med-MLLMs are recommended in their papers. Sometimes GPT-4V may refuse to generate response. For example, GPT-4V generated `I'm sorry, but I cannot provide the service of generating radiology reports. If you have any other questions or need information on a different topic, feel free to ask.'.

\begin{table*}[t]
\centering
\caption{Comparison of different MLLMs. The parameters of GPT-4V, GPT-4o, Gemini, and Claude 3.5 Sonnet are estimated to be greater than 100B.}
\begin{tabular}{>{\arraybackslash}p{3cm}|
			>{\arraybackslash}p{3cm}
            >{\arraybackslash}p{3cm}
            >{\arraybackslash}p{2cm}}
% \begin{tabular}{>{\arraybackslash}p{1.5cm}|
% 			>{\arraybackslash}p{2cm}
%             >{\arraybackslash}p{1cm}
%             >{\arraybackslash}p{1.3cm}}
%\begin{tabular}{c|ccc}
%\begin{tabular}{>{\arraybackslash}p{3cm}|
%			>{\arraybackslash}p{3cm}
%            >{\arraybackslash}p{2cm}
%            >{\arraybackslash}p{2cm}}

\hline
MLLMs & Vision Encoder & LLM & Parameters \\ \hline
GPT-4V & / & / & $\textgreater$100B \\ 
GPT-4o & / & / & $\textgreater$100B \\ 
Gemini & / & / & $\textgreater$100B \\ 
Claude 3.5 Sonnet & / & / & $\textgreater$100B \\ 
Med-Flamingo & ViT-L/14 & LLaMA & 8.3B \\ 
RadFM & ViT-3D & LLaMA & 14B \\ 
XrayGPT & MedClip & Vicuna & 7B \\ 
CheXagent & EVA-CLIP & Mistral & 7B \\  \hline
BiomedGPT-B & BERT & GPT-3 & 182M \\  \hline
\end{tabular}
\label{tab:llm}
\end{table*}

\begin{table*}[t]
\renewcommand\arraystretch{1.2}
\centering
\caption{The prompt we used to test different MLLMs.}
\begin{tabular}{>{\arraybackslash}p{3cm}| % Column 3 width with auto line break
			>{\arraybackslash}p{13.2cm}}
   %          >{\arraybackslash}p{1.5cm}| % Column 3 width with auto line break
			% >{\arraybackslash}p{5.2cm}}
%\begin{tabular}{>{\arraybackslash}p{3cm}| % Column 3 width with auto line break
%			>{\arraybackslash}p{9.1cm}}
%\begin{tabular}{c|c}
\hline
Med-MLLMs & Prompt \\ \hline
GPT-4v, GPT-4o, Gemini, and Claude 3.5 Sonnet& You are a professional doctor. I will give you a question and one or two images. Please utilize the image given to answer the question as a medical expert would. You should only give the answer and no reason or other information.  Question: $\textless$question$\textgreater$  Image: $\textless$image$\textgreater$. \\ \hline
Med-Flamingo & You are a helpful medical assistant. Please answer the following question. \\ \hline
RadFM & $\textless$image$\textgreater$ [image\_tokens] $\textless$/image$\textgreater$ [question] \\ \hline
XrayGPT & You are an experienced Doctor, give the following medical scan: $\textless$Img$\textgreater$ImageContent$\textless$/Img$\textgreater$ You will be able to see the medical scan once I provide it to you. Please answer my questions. \\ \hline
CheXagent and BiomedGPT-B & You are a professional doctor. I will give you a question and one or two images. Please utilize the image given to answer the question as a medical expert would. \\ \hline

\end{tabular}
\label{tab:prompt}
\end{table*}

\iffalse
\begin{table*}[h!]
\centering
	\caption{\textbf{The Analysis of Human Doctor in Capacities.}}
	\begin{tabular}{p{0.15\linewidth}|p{0.06\linewidth}p{0.06\linewidth}p{0.07\linewidth}|p{0.06\linewidth}p{0.06\linewidth}|p{0.06\linewidth}p{0.06\linewidth}|p{0.06\linewidth}}
		\toprule
  \hline
		\multirow{2}{*}{Human Doctor} & \multicolumn{3}{c|}{Perception} & \multicolumn{2}{c|}{Diagnosis} & \multicolumn{2}{c|}{Planning} & \multirow{2}{*}{Avg} \\ \cline{2-8} 
		& Anato & Attr & SpaQua & DisIde & Stag & Prog & Treat & \\ \hline
      Doctor1 & 0.541 & 0.554 & 0.567 & 0.554 & 0.567 & 0.554 & 0.567 & 0.564 \\ \hline
      Doctor2  & 0.460 & 0.469 & 0.490 & 0.483 & 0.490 & 0.483 & 0.490 & 0.498 \\ \hline
    Doctor3 & 0.542 & 0.502 & 0.512 & 0.502 & 0.512 & 0.502 & 0.512 & 0.538 \\ \hline
    Doctor4 & 0.545 & 0.536 & 0.550 & 0.543 & 0.550 & 0.543 & 0.550 & 0.536 \\ \hline
    
    Doctor5 & 0.420 & 0.545 & 0.548 & 0.545 & 0.548 & 0.545 & 0.548 & 0.465 \\ \hline

		\hline
		
		\bottomrule
	\end{tabular}
	\label{table:human_accuracy_capacity}
	
 \vspace{-0.3cm}
\end{table*}
\fi

% chatgpt evaluation for specialties

\begin{table*}[t]
	%\centering
	\caption{\textbf{Accuracy of models across different specialties.}}
	\begin{tabular}{p{0.2\linewidth}|p{0.07\linewidth}p{0.07\linewidth}p{0.07\linewidth}p{0.07\linewidth}p{0.07\linewidth}p{0.07\linewidth}p{0.07\linewidth}p{0.07\linewidth}}
		\toprule
		Model & Card & Derm & Endo & Gastro & GenSurg & Hem & Immun & Neurol \\
		\hline
            Doctor1 &     0.636 & 0.711 & 0.500 & 0.542 & 0.500 & 0.600 & 0.483 & 0.450 \\
            Doctor2 &     0.714 & 0.400 & 0.462 & 0.647 & 0.450 & 0.500 & 0.379 & 0.560 \\
            Doctor3 &     0.600 & 0.375 & 0.800 & 0.571 & 0.550 & 0.400 & 0.481 & 0.350 \\
            Meta-Doctor & 0.674 & 0.662 & 0.486 & 0.764 & 0.638 & 0.627 & 0.557 & 0.625 \\
		\hline
            GPT-4V & 0.571 & 0.627 & 0.440 & 0.598 & 0.508 & 0.454 & 0.633 & 0.598 \\ 
            GPT-4o & 0.561  & 0.609  & 0.560  & 0.619  & 0.292  & 0.505  & 0.733  & 0.611 \\
            Gemini & 0.390 & 0.361 & 0.400 & 0.459 & 0.292 & 0.399 & 0.500 & 0.463 \\ 
            CheXagent & 0.322 & 0.308 & 0.240 & 0.314 & 0.369 & 0.300 & 0.167 & 0.236 \\ 
            RadFM & 0.346 & 0.201 & 0.120 & 0.340 & 0.515 & 0.267 & 0.133 & 0.319 \\ 
            Med-Flamingo & 0.327 & 0.237 & 0.080 & 0.309 & 0.238 & 0.289 & 0.200 & 0.266 \\ 
            XrayGPT & 0.205 & 0.201 & 0.040 & 0.134 & 0.115 & 0.147 & 0.167 & 0.131 \\ 
            
            Claude3.5Sonnet & 0.229  & 0.095  & 0.040  & 0.268  & 0.331  & 0.040  & 0.000  & 0.240 \\
            % BiomedGPT-S & 0.068  & 0.053  & 0.000  & 0.077  & 0.215  & 0.106  & 0.000  & 0.004 \\
            % BiomedGPT-M & 0.088  & 0.089  & 0.000  & 0.119  & 0.192  & 0.143  & 0.000  & 0.039 \\
            BiomedGPT-B & 0.127  & 0.136  & 0.000  & 0.098  & 0.192  & 0.103  & 0.000  & 0.035 \\
		\bottomrule
	\end{tabular}
	\vspace{0.3 em}\\
	\begin{tabular}{p{0.2\linewidth}|p{0.06\linewidth}p{0.06\linewidth}p{0.06\linewidth}p{0.06\linewidth}p{0.06\linewidth}p{0.06\linewidth}p{0.06\linewidth}|p{0.06\linewidth}p{0.06\linewidth}} 
		\toprule
		Model & ObsGyn & Ophth & Orth & Oto & Path & Pulm & Urol & Avg & Var \\
		\hline
		Doctor1 &     0.455 & 0.800 & 0.423 & 0.800 & 0.474 & 0.500 & 0.636 & 0.574 & 0.014 \\
            Doctor2 &     0.500 & 0.846 & 0.500 & 0.286 & 0.333 & 0.500 & 0.750 & 0.538 & 0.023 \\
            Doctor3 &     0.750 & 0.737 & 0.500 & 0.500 & 0.400 & 0.600 & 0.722 & 0.578 & 0.020 \\
            Meta-Doctor & 0.667 & 0.756 & 0.600 & 0.448 & 0.581 & 0.689 & 0.700 & 0.641 & 0.007 \\
		\hline
            GPT-4V & 0.600 & 0.507 & 0.726 & 0.286 & 0.481 & 0.536 & 0.497 & 0.543 & 0.010 \\
            GPT-4o & 0.660  & 0.267  & 0.700  & 0.250  & 0.291  & 0.520  & 0.337  & 0.477  & 0.026 \\
            Gemini & 0.450 & 0.496 & 0.484 & 0.143 & 0.400 & 0.480 & 0.325 & 0.428 & 0.008 \\
            CheXagent & 0.260 & 0.437 & 0.326 & 0.107 & 0.326 & 0.440 & 0.313 & 0.343 & 0.007 \\
            RadFM & 0.240 & 0.299 & 0.321 & 0.214 & 0.314 & 0.291 & 0.282 & 0.302 & 0.009 \\
            Med-Flamingo & 0.260 & 0.425 & 0.274 & 0.214 & 0.286 & 0.347 & 0.215 & 0.302 & 0.006 \\
            XrayGPT & 0.120 & 0.214 & 0.158 & 0.107 & 0.128 & 0.113 & 0.141 & 0.148 & 0.002 \\
            
            Claude3.5Sonnet & 0.060  & 0.053  & 0.132  & 0.214  & 0.067  & 0.158  & 0.110  & 0.136  & 0.009 \\
            % BiomedGPT-S & 0.080  & 0.041  & 0.021  & 0.071  & 0.114  & 0.098  & 0.092  & 0.079  & 0.003 \\
            % BiomedGPT-M & 0.030  & 0.094  & 0.005  & 0.000  & 0.067  & 0.076  & 0.031  & 0.079  & 0.003 \\
            BiomedGPT-B & 0.070  & 0.126  & 0.016  & 0.036  & 0.072  & 0.104  & 0.018  & 0.090  & 0.003 \\
            
		\bottomrule
	\end{tabular}
	\label{table:models_accuracy}
	
\end{table*}

\section{Evaluation Metrics.} 
\label{appendix:metrics}
% \ourbenchmark~includes a range of question types: multiple choice, yes/no, open-ended questions, and report generation tasks. We adopt accuracy as metric for multiple-choice questions and yes/no questions. Moreover, open-ended questions demand a more subtle assessment approach; here, GPT is utilized to measure the precision of the textual responses. We use the prompt as following to judge if the answer is correct or not: \textit{You are an AI assistant who will help me evaluate responses given the questions and the correct answers. To assess a response, you should provide a single integer score like 0 or 1. A score of 0 indicates that the response is entirely different from answers. A score of 1 indicates that the response aligns perfectly with the answer or is correct for the given question and answer.
% [Question]
% [Answer] 
% [Response] 
% Your mark:
% }
% For the evaluation of report generation, the ROUGE-L scoring system is employed to determine the extent to which the models' generated texts align with the gold-standard reports\wt{~\citep{wu2023generalist,chen2024chexagent}}. For the overall accuracy calculation, we consider all questions in the benchmark apart from report generation. The accuracy is computed as the ratio of questions answered correctly to the total number of applicable questions.

\ourbenchmark~comprises a variety of question types, including multiple-choice, yes/no, open-ended questions, and report generation tasks. For multiple-choice and yes/no questions, we use accuracy as the evaluation metric. Open-ended questions require a more nuanced assessment, so we employ GPT to evaluate the precision of the textual responses. The following prompt is used to determine the correctness of an answer:

\textit{
You are an AI assistant who will help me evaluate responses given the questions and the correct answers. To assess a response, you should provide a single integer score like 0 or 1. A score of 0 indicates that the response is entirely different from answers. A score of 1 indicates that the response aligns perfectly with the answer or is correct for the given question and answer.
[Question]
[Answer] 
[Response] 
Your mark:
}
For report generation tasks, we utilize the ROUGE-L scoring system to evaluate how closely the generated texts match the gold-standard reports \wt{~\citep{wu2023generalist,chen2024chexagent}}. When calculating the overall accuracy, we include all question types except for report generation. The overall accuracy is calculated as the ratio of correctly answered questions to the total number of applicable questions in the benchmark.
\end{document}